\documentclass{article}

 \usepackage[main, final]{neurips_2026}

\usepackage{natbib}
\usepackage{enumitem}
\usepackage[utf8]{inputenc} 
\usepackage[T1]{fontenc}    
\usepackage{hyperref}
\hypersetup{
    colorlinks=true,
    linkcolor=blue,       
    citecolor=cyan,       
    urlcolor=blue         
}
\usepackage{url}            
\usepackage{booktabs}       
\usepackage{amsfonts}       
\usepackage{nicefrac}       
\usepackage{microtype}      
\usepackage{xcolor}         

\usepackage{graphicx}
\usepackage{wrapfig}
\usepackage{multirow}
\usepackage{makecell}
\usepackage{gensymb}
\usepackage{amsmath}
\usepackage{amsthm}
\usepackage{mathtools}
\usepackage{caption}
\usepackage{adjustbox}

\newtheorem{theorem}{Theorem}[section]
\newtheorem{lemma}[theorem]{Lemma}

\usepackage{tcolorbox}
\tcbuselibrary{skins, breakable} 
\newtheorem{assumption}{Assumption}[section]
\newtheorem{definition}{Definition}[section]
\usepackage{algorithm}
\usepackage{algorithmic}
\usepackage{amssymb}

\usepackage[table]{xcolor}
\usepackage{fontawesome5}

\title{Rethinking Sequential Locate-Then-Edit:\\ Optimality and Stability}

\author{%
  Bingqing Liu$^{1}$\quad 
  Wei Liu$^{1,\ast}$ \quad Zhiying Deng$^{2}$ \quad Jun Wang$^{3}$ \quad
  \quad Yuhua Li$^{1}$ \quad Ruixuan Li$^{1}$ \\[0.5em]
  $^{1}$School of Computer Science and Technology, Huazhong University of Science and Technology\\
   $^{2}$ Laboratory for Artificial Intelligence and New Forms of Education, CCNU \quad $^{3}$iWudao Tech  \\
  \texttt{\{liubingqing24, idc\_lw, idcliyuhua, rxli\}@hust.edu.cn}\\
  \texttt{zhiyingdzy@ccnu.edu.cn, jwang@iwudao.tech}\\
  $^*$Corresponding authors.\quad \faGithub\ \href{https://github.com/lbq8942/BESE}{Code}
}

\begin{document}

\maketitle

\begin{abstract}
The \textit{Locate-Then-Edit} framework  enables efficient knowledge updates in large language models without requiring full retraining. Recent research has increasingly focused on \textbf{sequential editing}, where knowledge arrives continuously and the model is updated incrementally. However, existing sequential methods commonly suffer from catastrophic forgetting and model collapse after only a few hundred edits. In contrast, \textbf{batch editing}—which jointly optimizes all edits and yields the optimal solution under a specified editing objective—can support tens of thousands of edits with high edit efficacy while preserving the model’s general capabilities, resulting in a substantial performance gap between the two editing paradigms. In this paper, we demonstrate that this gap can be closed via a \textbf{batch-equivalent sequential editing} approach. Moreover, we investigate the underlying mechanism of model degradation in existing sequential methods by analyzing the convergence behavior of weight perturbations as edits accumulate. Our analysis reveals that current methods typically exhibit linear or logarithmic growth in perturbation, whereas the proposed batch-equivalent method achieves time-invariant growth, thereby theoretically guaranteeing long-term stability and preserving the integrity of pretrained knowledge throughout massive-scale sequential editing.
\end{abstract}

\section{Introduction}
\label{sec:introduction}

Updating the knowledge encoded in the parameters of Large Language Models (LLMs) is essential for refreshing outdated facts and integrating domain-specific information to enable effective customization \citep{sanu2024limitations}. However, retraining LLMs from scratch is often prohibitively expensive due to the immense computational cost and time required. While fine-tuning offers a more practical alternative, it frequently suffers from catastrophic forgetting of previously acquired knowledge \citep{luo2025empirical}. To address these challenges, knowledge editing (KE) methods \citep{wang2024knowledge} have emerged as promising approaches that allow precise, efficient, and localized modifications to specific factual associations within LLMs without compromising model's behavior on unrelated knowledge.

Among KE paradigms, the \textit{locate-then-edit} framework has garnered significant attention \citep{meng2022locating}. Early efforts within this framework primarily focused on \textit{batch editing} \citep{mengmass,gupta2024unified}, in which all edits are applied and optimized jointly, yielding superior performance under large-scale editing scenarios. However, real-world knowledge evolves continuously, demanding timely updates for LLMs to remain accurate and relevant \citep{jiang2025neuron}. This practical demand has made sequential (also called lifelong or continual) knowledge editing a significant and growing focus of research \citep{gu2024model,li2024can,li2025adaedit,liubetaedit,liu2026gedit}.

\begin{wrapfigure}[21]{r}{0.5\linewidth}
  \centering
  \includegraphics[width=0.95\linewidth]{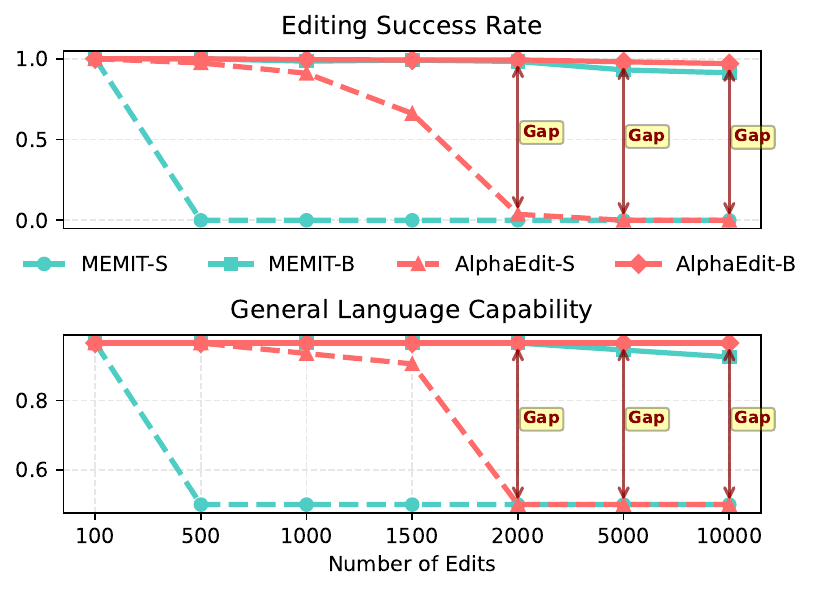}
\caption{
  Performance gap between sequential editing (-S) and batch editing (-B) on the Counterfact dataset using LLaMA3-8B-Instruct. \textbf{Top}: Editing success rate. \textbf{Bottom}: General language capability (accuracy on SST sentiment classification from GLUE benchmark).
}
  \label{fig:intro}
\end{wrapfigure}

In sequential editing, edits arrive one after another, and the model must be updated incrementally. This paradigm introduces two key challenges: repeated editing can lead to (1) forgetting of previously edited knowledge and (2) degradation of the model’s pre-trained general capabilities, often termed \textit{model collapse} in severe cases \citep{gupta2024rebuilding,fangalphaedit}. The former compromises editing effectiveness, while the latter is especially catastrophic: once foundational language understanding deteriorates, the model becomes fundamentally unusable, which in turn invalidates any subsequent editing attempts \citep{gupta2024model}. Figure \ref{fig:intro} illustrates the editing performance of state-of-the-art approaches under 10,000 sequential edits. While the \textit{history-regularized} editor AlphaEdit \citep{fangalphaedit} significantly outperforms the \textit{history-agnostic} MEMIT \citep{mengmass}, both methods  fail after only 2,000 edits, suffering sharp declines in both editing success rate and general language capability. In stark contrast, batch editing remains effective even after 10,000 edits, revealing a substantial performance gap between batch and sequential approaches.

This observation raises two fundamental questions:  
(1) Is the gap between batch and sequential editing inevitable? Can we design a sequential method that \textit{exactly closes} this gap?  
(2) Why does \textit{batch editing} better preserve the model's general capabilities than \textit{history-regularized} editing, and why does history-regularized editing outperform \textit{history-agnostic} editing? What underlying mechanism drives the progressive degradation across these strategies?

In this paper, within the \textit{locate-then-edit} framework, we address both questions. First, we propose a sequential editing algorithm that is \textit{exactly equivalent} to batch editing: at every editing step, the model produced by our method is identical to the one obtained by applying all edits up to that point via a single batch optimization, thereby attaining the optimal solution under the specified editing objective. Second, we conduct a convergence analysis of existing sequential editing methods by characterizing the temporal evolution of model weights. Our analysis reveals a fundamental distinction in weight perturbation dynamics: our method (which matches batch editing) exhibits \textit{time-invariant} perturbations; in contrast, history-aware methods suffer from \textit{logarithmic growth}, while history-agnostic methods incur \textit{linear growth} in perturbation magnitude. This theoretical insight explains the empirical model degradation trends in Figure \ref{fig:intro} and establishes a clear connection between update dynamics and long-term editing stability.

\section{Related Work}

Model editing aims to update specific knowledge in large language models while preserving unrelated facts and general capabilities. Approaches are broadly categorized by whether they follow the \textit{locate-then-edit} (LTE) paradigm.

\subsection{Locate-Then-Edit (LTE) Model Editing}
LTE operates in two stages: first, it identifies model components—such as activations or parameter subsets—causally responsible for a target fact, often via neuron-level attribution or causal tracing   \citep{geva2021transformer,yu2024neuron,zhou2025editing,guomechanistic,jiang2025neuron,liu2025edit}; second, it applies targeted modifications. ROME \citep{meng2022locating} pioneered this by editing feed-forward network (FFN) parameters for single-fact updates. Later works extended ROME to batch settings: MEMIT \citep{mengmass}, PMET \citep{li2024pmet}, and EMMET \citep{gupta2024unified} support thousands of concurrent edits. To better preserve pre-trained knowledge, methods based on null-space constraints and adaptive regularization have been introduced \citep{fangalphaedit,liubetaedit,liu2026gedit}. Subsequent research has explored complex scenarios including long-form editing \citep{jianganyedit,zeng2025docmedit,li2025scedit}, multilingual and multimodal updating \citep{sun2025mitigating,wang2025ace}, and multi-hop reasoning \citep{zhong2023mquake,zhang2025locatethenedit,thedewikibigedit}.

A key challenge in LTE is \textit{sequential editing}, where repeated updates degrade performance, e.g., eroding basic language understanding capability \citep{gu2024model}. Two main strategies address this: (1) post-editing weight regularization, which constrains parameters or their spectral properties \citep{gu2024model,maperturbation,li2025adaedit,liu2025edit,qiao2025wasserstein,liu2026energyregularized}; and (2) history-aware regularization that leverages past edits for stability \citep{fangalphaedit,guotowards,fang2025hippocampal,liubetaedit}.

\subsection{Non-LTE Model Editing}

Non-LTE methods can be categorized according to whether
they modify the original model parameters. Parameter-preserving methods keep the original model parameters unchanged
and incorporate edits through auxiliary mechanisms.
Examples include external memory modules
\citep{mitchell2022memory}, neuron modulation modules
\citep{dong2022calibrating,huangtransformer},
external knowledge bases
\citep{hartvigsen2023aging,fei2025neuraldb},
low-rank adapters \citep{yu2024melo}, and side networks
\citep{wang2024wise}. Other methods intervene at inference
time, for example, by augmenting prompts with retrieved
information to guide model outputs without modifying
model parameters
\citep{madaan2022memory,zheng2023can,wang2024retrieval,gu2024pokemqa,bi2025decoding,wang2025decoupling}. Parameter-modifying methods edit knowledge by directly modifying the
model's internal parameters. One line of work adopts direct fine-tuning
\citep{gangadhar2024model,wang2024roselora,liumitigating,liuunlocking,piao3},
while another employs meta-learning to generate weight updates via hypernetworks, as in
MEND \citep{mitchellfast}, MALMEN \citep{tanmassive},
InstructEdit \citep{zhang2024instructedit}, LKS
\citep{liu2025latent}, and RLEdit \citep{lireinforced}.
\section{Background}
\label{sec:background}

\paragraph{Knowledge Location} Our scope is the \textit{locate-then-edit} (LTE) paradigm. LTE first identifies internal model elements that most influence a factual output using causal tracing, gradient sensitivity, or neuron attribution \citep{yu2024neuron,zhou2025editing,guomechanistic,jiang2025neuron,liu2025edit}. It then applies targeted modifications. In Transformers, factual knowledge is often stored in feed-forward networks (FFNs) as key-value associative memories \citep{geva2021transformer}. Within an FFN, the first linear layer computes a key, and the second retrieves the value. Early-layer FFNs initiate fact retrieval \citep{meng2022locating}. Thus, mainstream LTE methods edit the second linear layer's weights in early FFN blocks to insert or alter knowledge \citep{mengmass}.

\paragraph{Knowledge Representation} For a factual triple $(s, a, o)$ — e.g., (\textit{United States}, \textit{President}, \textit{Donald Trump}) — model editing aims to make the model generate target answer $o$ when prompted with $\textit{prompt}(s, a)$ (e.g., ``The president of the United States is''). Let $\mathbf{W}_0$ be the weight matrix identified as storing this fact, mapping input key $k$ to output value $v$. The key $k$ is the last-token activation of subject $s$ (e.g., \textit{States}), empirically most effective for editing \citep{meng2022locating}. To steer the model toward $o$, the target value $v$ is optimized via gradient descent to maximize the likelihood of generating $o$ given $\textit{prompt}(s, a)$.

\paragraph{Knowledge Editing} Suppose we are provided with $t$ key–value edit pairs $(k_i, v_i)_{i=1}^t$. The goal of knowledge editing is to incorporate new knowledge while preserving pre-trained knowledge. Let $\mathbf{K}_0 = [k_1^{\text{pre}}, \dots, k_n^{\text{pre}}]$ and $\mathbf{V}_0 = [v_1^{\text{pre}}, \dots, v_n^{\text{pre}}]$ denote the matrices of pre-trained keys and values from the initial weight $\mathbf{W}_0$. The objective is to find a perturbation $\Delta$ such that $\mathbf{W}_0 + \Delta$ correctly maps new pairs $(k_i, v_i)_{i=1}^t$ while minimally altering behavior on $(\mathbf{K}_0,\mathbf{V}_0)$. This yields the following constrained optimization problem \citep{meng2022locating}:
\begin{equation}
\begin{aligned}
\Delta^* = \arg\min_{\Delta} \; \left\| (\mathbf{W}_0 + \Delta) \mathbf{K}_0 - \mathbf{V}_0 \right\|_F^2, \quad\text{s.t.} \; (\mathbf{W}_0 + \Delta) k_i = v_i, \quad \forall\, i = 1, \dots, t.
\label{eq:opt}
\end{aligned}
\end{equation}

\section{Method}
\label{sec:method}
The formulation in Eq. \ref{eq:opt} defines a batch editing scheme that optimizes all $t$ edits simultaneously. While effective, it fails to handle edits that arrive sequentially in practice. Furthermore, storing all key–value pairs incurs linear memory growth in $t$, which becomes prohibitive at scale. These limitations motivate sequential editing.
\subsection{Problem Formulation}
We first formalize the sequential editing problem \citep{maperturbation,lireinforced}, which aims to incrementally update the model weights as new edit requests arrive.
\begin{definition}[\textbf{Sequential Editing}]
Consider a sequence of edit requests  $ (k_i, v_i)_{i=1}^t $ , where each  $ k_i$  is an input key and  $ v_i$  is the desired output value. A sequential editing method is defined by an update operator  $ f $  that recursively updates the model weight  starting from the initial weight  $ \mathbf{W}_0 $: 
$$
\mathbf{W}_{j+1} = f(\mathbf{W}_j, k_{j+1}, v_{j+1}),
$$
\end{definition}
Before designing such an operator $f$, we must first define what constitutes an \textit{optimal} edit. Specifically, given the first $j$ edit requests, what is the ideal weight matrix $\mathbf{W}_j^*$? Only with this target clearly defined can we design a sequential algorithm $f$ whose output $\mathbf{W}_j$ either approximates or, ideally, exactly matches the optimal solution $\mathbf{W}_j^*$. 

At its core, \textit{locate-then-edit} can be viewed as a key-value remapping problem: given a set of new key-value pairs to insert and pretrained key-value pairs to preserve, an editing objective specifies the criteria for achieving this remapping. Different choices of objectives lead to different notions of optimality. Prior work by \citet{guotowards} defines the optimal edit in alignment with ROME \citep{meng2022locating} and EMMET \citep{gupta2024unified}, which impose equality constraints on the new key--value pairs. In contrast, we adopt the formulation from MEMIT \citep{mengmass} as our optimal editing objective, which casts editing as a regularized least-squares problem.

\begin{definition}[\textbf{Optimal Editing Objective}]
\label{def:opt}
Let $\mathbf{K}_{[1:j]} = [k_1, \dots, k_j]$ and
$\mathbf{V}_{[1:j]} = [v_1, \dots, v_j]$ denote the new
key-value pairs to be inserted, and let
$\mathbf{K}_0 = [k_1^{\text{pre}}, \dots, k_n^{\text{pre}}]$
and
$\mathbf{V}_0 = [v_1^{\text{pre}}, \dots, v_n^{\text{pre}}]$
represent the pre-trained key-value pairs to be preserved.
We adopt the following regularized least-squares objective as the optimal editing objective:
\begin{equation}
\begin{split}
\mathcal{L}_j(\Delta) =
\lambda_1 \big\|(\mathbf{W}_0+\Delta)
\mathbf{K}_{[1:j]}-\mathbf{V}_{[1:j]}\big\|_F^2
+\lambda_2 \big\|(\mathbf{W}_0+\Delta)
\mathbf{K}_0-\mathbf{V}_0\big\|_F^2.
\end{split}
\label{eq:opt_memit}
\end{equation}
\end{definition}

We augment the editing objective (Eq. \ref{eq:opt_memit}) with two  coefficients  $ \lambda_1 $  and  $ \lambda_2 $  to obtain a more general formulation. Specifically, it unifies major existing \emph{locate-then-edit} (LTE) methods as limiting cases: as \(\lambda_1 \to \infty\), the solution enforces exact satisfaction of the new edits, recovering methods such as ROME and EMMET; conversely, as \(\lambda_2 \to \infty\), the update is constrained to the null space of \(\mathbf{K}_0\), prioritizing preservation of pre-trained knowledge, as in AlphaEdit \citep{fangalphaedit}. A  detailed discussion is provided in Appendix \ref{app:unify}. 

Minimizing the objective in Eq.~\ref{eq:opt_memit} yields the optimal weight update \citep{mengmass}:
\begin{equation}
\begin{split}
\Delta_j^* = \arg\min_{\Delta} \mathcal{L}_j(\Delta)
= \lambda_1\mathbf{R}_{[1:j]}\mathbf{K}_{[1:j]}^\top
\left(
\lambda_1\mathbf{K}_{[1:j]}\mathbf{K}_{[1:j]}^\top
+\lambda_2\mathbf{K}_0\mathbf{K}_0^\top
\right)^{-1},
\end{split}
\label{eq:solution}
\end{equation}
where $\mathbf{R}_{[1:j]} = [r_1,\dots,r_j]
= \mathbf{V}_{[1:j]}-\mathbf{W}_0\mathbf{K}_{[1:j]}$
denotes the matrix of residuals. Accordingly, the optimal weight matrix is given by $\mathbf{W}_j^*=\mathbf{W}_0+\Delta_j^*$. Thus, under our definition, an optimal sequential operator $f^*$ should yield the same weight update as batch editing, which optimizes all edits jointly.

\subsection{Batch-Equivalent Sequential Editing}

For the first edit ($j=1$), sequential editing trivially coincides with batch editing. Our goal is to extend this property to arbitrary $j$: can we derive a sequential update operator $f^*$ such that applying it sequentially yields the same result as batch editing at every step? Formally, we seek $f^*$ satisfying
\[
\mathbf{W}_{j+1}^* = f^*(\mathbf{W}_j^*, k_{j+1}, v_{j+1}), \qquad \forall j \ge 1,
\]
where $\mathbf{W}_j^*$ denotes the  batch solution for the first $j$ edits. When such an operator exists, it is by Definition~\ref{def:opt} optimal and called \textit{batch-equivalent}.

To derive such an operator, we start from the closed-form batch solution in Eq.~\ref{eq:solution}. Define
\begin{equation}
    A_j := \lambda_1\sum_{i=1}^{j} k_i k_i^\top + \lambda_2\mathbf{K}_0 \mathbf{K}_0^\top,
    \quad
    B_j := \lambda_1 \sum_{i=1}^{j} r_i k_i^\top.
    \label{eq:solj}
\end{equation}
By Eq.~\ref{eq:solution}, the optimal weight update for the first $j$ edits is $\Delta_j^* = B_j A_j^{-1}$, with $\mathbf{W}_j^* = \mathbf{W}_0 + \Delta_j^*$. We aim to find a sequential operator $f^*$ that connects $\mathbf{W}_{j+1}^*$ and $\mathbf{W}_j^*$. The following theorem provides such an operator.

\begin{theorem}[\textbf{Batch-Equivalent Sequential Update}]
\label{thm:batch_equivalent_memit}
Let $\mathbf{W}_j^*$ denote the optimal weight matrix for the first $j$ edits. Then the optimum $\mathbf{W}_{j+1}^*$ at the next step can be obtained from $\mathbf{W}_j^*$ via
\begin{equation}
\mathbf{W}_{j+1}^* = \mathbf{W}_j^* \big( \mathbf{I} - \lambda_1 k_{j+1} k_{j+1}^\top A_{j+1}^{-1} \big) + \lambda_1 \big( r_{j+1} + \mathbf{W}_0 k_{j+1} \big) k_{j+1}^\top A_{j+1}^{-1},
\label{eq:optimal_update}
\end{equation}
where  the right-hand side defines the optimal sequential operator $f^*$.  
\end{theorem}

Here the Gram matrix $A_{j+1}$ can be computed incrementally via $A_{j+1} = A_j +\lambda_1 k_{j+1}k_{j+1}^\top$, with $A_j$ cached from the $j$-th step. Starting from $j=1$, the operator $f^*$ sequentially updates both the Gram matrix $A_j$ and the weight matrix $\mathbf{W}_j^*$, thereby matching the batch solution at every editing step.  Its proof, which relies on the Sherman–Morrison–Woodbury (SMW) identity \citep{golub2013matrix} and straightforward algebra, is deferred to Appendix~\ref{app:bese}. 

Theorem~\ref{thm:batch_equivalent_memit} gives the optimal sequential operator that matches the MEMIT-style editing objectives. Similarly, optimal sequential operators can be derived for other LTE methods under their respective editing objectives. For AlphaEdit, for instance, the operator takes the following form:
\[
\mathbf{W}_{j+1} = \mathbf{W}_j \big( \mathbf{I} - k_{j+1} k_{j+1}^\top \mathbf{P} A_{j+1}^{-1} \big) + \big( r_{j+1} + \mathbf{W}_0 k_{j+1} \big) k_{j+1}^\top \mathbf{P} A_{j+1}^{-1},
\]
where $\mathbf{P}$ is the projection matrix.  The derivation is also deferred to Appendix~\ref{app:bese}.

In Appendix~\ref{app:bese}, we show that although
Theorem~\ref{thm:batch_equivalent_memit} assumes one edit
at a time, it naturally generalizes to mini-batch sequential
editing. We also discuss how the proposed sequential operator
$f^*$ degenerates to existing sequential methods when certain
components are removed. Note that although $f^*$ is theoretically
equivalent to batch editing, finite-precision recursive updates
may introduce numerical errors that propagate across editing
steps, leading to deviations from the batch solution. We 
discuss these numerical discrepancies and practical implementation
considerations in Appendix~\ref{app:precision_impl}.

\section{Stability Analysis}
\label{sec:sta}
In prior work, the failure of sequential editing is typically observed through surface-level symptoms—such as degradation on general language understanding benchmarks (e.g., GLUE) \citep{gu2024model,gupta2024model,maperturbation,parksuit} or shifts in the distribution of hidden representations in Transformers \citep{fangalphaedit,li2025adaedit,lirethinking,liu2025edit}. However, these observations do not address the root cause: since the LTE method directly modifies model parameters, the fundamental issue lies in how the weight matrix  $\mathbf{W}_t$ evolves over  successive edits. While a few studies analyze weight dynamics, they are either limited to single-edit settings or rely on overly idealized assumptions \citep{maperturbation,fang2025hippocampal,liu2026energyregularized}; see Appendix \ref{app:sta} for a detailed discussion.

To systematically investigate the weight dynamics during sequential editing, we first introduce two mild conditions that are standard, and often implicitly satisfied, in current LTE frameworks:
\begin{assumption}[\textbf{Boundedness and Invertibility}]
\label{ass:well_posedness}
The following conditions hold throughout the sequential editing process:
\begin{enumerate}[label=(\roman*)]
    \item \textbf{Boundedness:} There exist constants $C_k > 0$ and $C_r > 0$ such that for all edit steps $t$, $\|k_t\|_2 \le C_k$ and $\|r_t\|_2 \le C_r$.
    \item \textbf{Invertibility:} There exists a constant $\alpha > 0$ such that $\mathbb{E}[k k^\top] \succeq \alpha \mathbf{I}$, where $k$ denotes any key vector from either the pre-trained knowledge bank or an incoming edit request. This ensures that the Gram matrix in Eq.~\eqref{eq:solution} remains well-conditioned during long-term editing, enabling stable inversion in LTE updates.
\end{enumerate}
\end{assumption}
Under Assumption~\ref{ass:well_posedness}, we present a unified characterization of the weight dynamics across existing  sequential LTE methods and our batch-equivalent approach. To this end, we first categorize existing approaches into three representative classes: history-agnostic sequential editing, history-regularized sequential editing, and post-hoc regularized sequential editing.

\paragraph{History-Agnostic Editing}
Methods in this class apply each edit independently, without accounting for prior edits \citep{meng2022locating,mengmass,gupta2024unified}. The update rule is:
\[
\mathbf{W}_t = \mathbf{W}_{t-1} + r_t k_t^\top (k_t k_t^\top + \mathbf{K}_0 \mathbf{K}_0^\top)^{-1}.
\]

\paragraph{History-Regularized Editing}
History-regularized methods incorporate past edits $\mathbf{K}_{[1:t-1]}$ into the regularization term to preserve previously learned key–value associations. Initially proposed by \citet{fangalphaedit,guotowards,fang2025hippocampal}, this mechanism has since become a \textit{de facto} prerequisite for achieving state-of-the-art performance in sequential editing \citep{lirethinking,liu2025edit,liu2026energyregularized,parksuit}. The update rule is given by:
\[
\mathbf{W}_t = \mathbf{W}_{t-1} + r_t k_t^\top \bigl( k_t k_t^\top + \mathbf{K}_{[1:t-1]} \mathbf{K}_{[1:t-1]}^\top + \mathbf{K}_0 \mathbf{K}_0^\top \bigr)^{-1}.
\]

\paragraph{Post-Hoc Regularized Editing}
Given a weight perturbation $\Delta$ computed by a third-party editing method, post-hoc regularization applies a regularization constraint. Examples include zeroing out small-magnitude entries \citep{gu2024model}, discarding singular components with small singular values \citep{li2025adaedit}, or projecting the perturbation onto a non-core subspace to preserve the original weight space \citep{liu2026energyregularized,parksuit}. Despite their differences in strategy, these approaches can be uniformly expressed by the following update rule:
\[
\mathbf{W}_t = \mathbf{W}_{t-1} + \operatorname{PR}(\Delta, \alpha),
\]
where $\alpha$ is a hyperparameter controlling the regularization strength. For instance, if the regularization operation zeroes out small entries, then $\alpha$ denotes the proportion of entries to zero out. In this case, $\alpha = 1$ yields $\operatorname{PR}(\Delta, \alpha) = \mathbf{0}$, and $\alpha = 0$ yields $\operatorname{PR}(\Delta, \alpha) = \Delta$.

Given these update rules, we now analyze their stability during long-term sequential editing by studying the evolution of the Frobenius norm $\|\mathbf{W}_t\|_F$. Since unbounded growth of this norm can lead to numerical instability or model collapse, we focus on deriving the upper bound of  $ \|\mathbf{W}_t\|_F $, which characterizes the worst-case magnitude of weight perturbations. (The lower bound is trivially zero for all sequential editing methods and thus less informative for stability analysis.) Under Assumption \ref{ass:well_posedness}, we establish the following guarantees for weight dynamics:
\begin{theorem}
\label{thm:norm_growth}
Let $\lambda_{\min}$ denote the smallest eigenvalue of $\mathbf{K}_0 \mathbf{K}_0^\top$. The Frobenius norm of $\mathbf{W}_t$ for the above classes of sequential LTE methods satisfies the following tight upper bounds:
\begin{align*}
\|\mathbf{W}_t^{\mathrm{HA}}\|_F &\le \|\mathbf{W}_0\|_F + \frac{C_r}{2\sqrt{\lambda_{\min}}} \cdot t, \\
\|\mathbf{W}_t^{\mathrm{HR}}\|_F &\le \|\mathbf{W}_0\|_F + \frac{C_r C_k}{\alpha}\,(1 + \log t), \\
\|\mathbf{W}_t^{\mathrm{BE}}\|_F &\le \|\mathbf{W}_0\|_F + \frac{C_r C_k}{\alpha}, \\
\|\mathbf{W}_t^{\mathrm{PR}}\|_F &\le \|\mathbf{W}_0\|_F + B_{\mathrm{base}}(t),
\end{align*}
where $\mathbf{W}_t^{\mathrm{HA}}$, $\mathbf{W}_t^{\mathrm{HR}}$, $\mathbf{W}_t^{\mathrm{BE}}$, and $\mathbf{W}_t^{\mathrm{PR}}$ denote the weights under history-agnostic, history-regularized, batch-equivalent, and post-hoc regularized sequential editing, respectively. $B_{\mathrm{base}}(t)$ denotes the corresponding upper bound of the underlying base editor (HA, HR, or BE).

\end{theorem}

Proofs are provided in Appendix~\ref{app:convergence}. Theorem~\ref{thm:norm_growth} characterizes the worst-case weight growth during sequential editing: $\|\mathbf{W}_t^{\mathrm{HA}}\|_F$ can grow linearly with $t$, $\|\mathbf{W}_t^{\mathrm{HR}}\|_F$ at most logarithmically, and $\|\mathbf{W}_t^{\mathrm{BE}}\|_F$ remains uniformly bounded—demonstrating the superior long-term editing stability of our approach. For post-hoc regularized editing ($\|\mathbf{W}_t^{\mathrm{PR}}\|_F$), the norm bound inherits that of its underlying base editor (HA, HR, or BE). Although these bounds are worst-case guarantees, our experiments (Section~\ref{sec:exp}) reveal a clear and consistent gap in the actual norm growth across these methods, closely matching the theoretical growth hierarchy.

\section{Experiments}
\label{sec:exp}

\begin{figure}[t]
  \centering
  \includegraphics[width=1\linewidth]{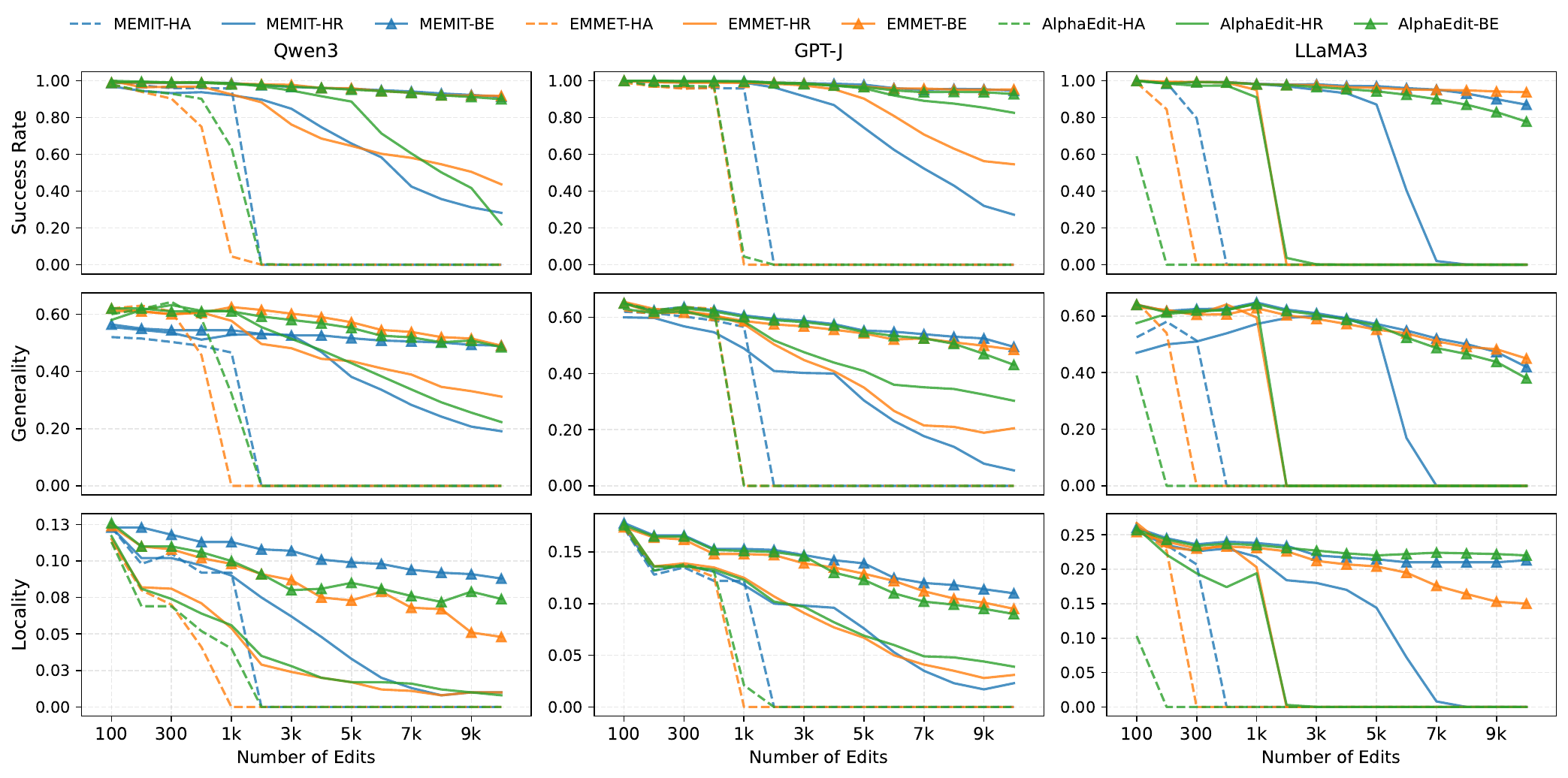}
\caption{
Editing performance (success rate, generality, locality) on the \textsc{CounterFact} dataset over 10,000 sequential edits. Columns (from left to right): Qwen3-8B, GPT-J, LLaMA3-8B-Instruct. Higher values indicate better performance for all metrics.}
  \label{fig:performance_cf}
\end{figure}

\subsection{Experimental Setup}

Our experimental design strictly follows the taxonomy introduced in Section~\ref{sec:sta}. We investigate how different sequential editing paradigms—history-agnostic, history-regularized, batch-equivalent, and post-hoc regularized—affect both editing performance and editing stability.

\paragraph{LLMs and Datasets.}
We evaluate our methods on four autoregressive language models:
GPT2-XL-1.5B, GPT-J-6B, Qwen3-8B, and LLaMA3-8B-Instruct. These models cover a range of model architectures and sizes.
All model checkpoints are publicly available through the
Hugging Face Hub.\footnote{\url{https://huggingface.co/models}}
Following prior work~\citep{fangalphaedit}, we perform edits
on layers $\{13,14,15,16,17\}$ for GPT2-XL,
layers $\{3,4,5,6,7,8\}$ for GPT-J, and layers
$\{4,5,6,7,8\}$ for Qwen3-8B and LLaMA3-8B-Instruct.
Results on GPT2-XL are deferred to Appendix~\ref{app:results}. Evaluations are conducted on two standard factual editing benchmarks: \textsc{CounterFact}~\citep{meng2022locating} and \textsc{ZsRE}~\citep{levy2017zero}. Throughout this paper, the edit sequence length is fixed at 10,000. We adopt the most challenging editing setting—applying only one edit per step (i.e., batch size = 1). This setup allows us to rigorously stress-test the model’s sequential editing performance over 10,000 update steps.

\paragraph{Base Editors and Post-Hoc Regularization Strategies.}
We build upon three foundational LTE editors: MEMIT~\citep{mengmass}, EMMET~\citep{gupta2024unified} (an extension of ROME~\citep{meng2022locating}), and AlphaEdit~\citep{fangalphaedit}. For each, we develop \textbf{history-agnostic} (MEMIT-HA, EMMET-HA, AlphaEdit-HA), \textbf{history-regularized} (MEMIT-HR, EMMET-HR, AlphaEdit-HR), and \textbf{batch-equivalent} (MEMIT-BE, EMMET-BE, AlphaEdit-BE) variants. Their update rules are detailed in Appendix~\ref{app:baseline_hahr}. For post-hoc regularization, we include \textsc{RECT}~\citep{gu2024model}, \textsc{PRUNE}~\citep{maperturbation}, \textsc{AdaEdit}~\citep{li2025adaedit}, and \textsc{SPHERE}~\citep{liu2026energyregularized}. Additional baselines (e.g., parameter-preserving methods) are discussed in Appendix~\ref{app:results}.

\subsection{Hierarchical Editing Performance}
\begin{table*}[t]
\centering
\caption{Detailed performance of MEMIT-BE on Qwen3-8B across
different numbers of sequential edits on CounterFact and ZsRE.
SR., Gen., and Loc. denote Success Rate, Generality, and Locality,
respectively. SST, MMLU, CoLA, and NLI report accuracy (\%) on
the corresponding general language understanding benchmarks.}
\label{tab:qwen_perf}
\setlength{\tabcolsep}{3pt}
\resizebox{0.95\textwidth}{!}{
\begin{tabular}{l ccc cccc |ccc cccc}
\toprule
\multirow{2}{*}{Edits} & \multicolumn{7}{c|}{CounterFact} & \multicolumn{7}{c}{ZsRE} \\
\cmidrule(lr){2-8} \cmidrule(lr){9-15}
 & SR. & Gen. & Loc. & SST & MMLU & CoLA & NLI
 & SR. & Gen. & Loc. & SST & MMLU & CoLA & NLI \\
\midrule
1k  & 98.8 & 54.4 & 11.3 & 83.0 & 72.5 & 71.5 & 83.5 
    & 94.8 & 89.1 & 38.2 & 84.0 & 73.5 & 72.0 & 82.5 \\
2k  & 97.6 & 53.1 & 10.8 & 83.0 & 72.5 & 70.5 & 84.0 
    & 94.5 & 88.0 & 38.1 & 84.5 & 75.5 & 72.0 & 82.5 \\
3k  & 96.7 & 52.6 & 10.7 & 83.5 & 72.5 & 71.5 & 83.0 
    & 93.9 & 88.4 & 37.6 & 81.0 & 75.0 & 72.0 & 81.5 \\
4k  & 96.1 & 52.6 & 10.1 & 83.5 & 72.0 & 71.5 & 83.0 
    & 93.6 & 87.2 & 37.6 & 81.5 & 75.5 & 71.0 & 81.5 \\
5k  & 95.7 & 51.6 & 9.9  & 84.0 & 72.5 & 72.0 & 83.0 
    & 93.4 & 86.7 & 37.8 & 82.5 & 74.5 & 73.5 & 82.5 \\
6k  & 94.9 & 50.8 & 9.8  & 83.5 & 70.5 & 69.0 & 82.0 
    & 92.5 & 86.1 & 37.5 & 83.5 & 74.5 & 75.0 & 81.5 \\
7k  & 94.2 & 50.5 & 9.4  & 83.5 & 72.5 & 70.0 & 81.5 
    & 91.9 & 85.0 & 37.3 & 81.0 & 73.5 & 75.5 & 80.0 \\
8k  & 93.1 & 50.1 & 9.2  & 82.0 & 71.0 & 70.0 & 82.0 
    & 91.3 & 84.8 & 37.5 & 81.0 & 74.5 & 74.5 & 81.0 \\
9k  & 92.3 & 49.3 & 9.1  & 80.5 & 69.5 & 72.5 & 81.5 
    & 91.2 & 84.7 & 37.3 & 81.5 & 75.0 & 74.0 & 83.0 \\
10k & 91.3 & 49.0 & 8.8  & 81.5 & 70.5 & 72.0 & 80.5 
    & 90.9 & 84.5 & 37.2 & 82.5 & 70.5 & 73.0 & 83.0 \\
\bottomrule
\end{tabular}
}
\end{table*}
\begin{table*}[t]
\centering
\caption{Layer-wise weight norms of MEMIT-BE on Qwen3-8B across different numbers of sequential edits on CounterFact and ZsRE.}
\label{tab:qwen_norm}
\setlength{\tabcolsep}{3pt}
\resizebox{0.95\textwidth}{!}{
\begin{tabular}{l ccccc ccccc}
\toprule
\multirow{2}{*}{Edits} & \multicolumn{5}{c}{CounterFact} & \multicolumn{5}{c}{ZsRE} \\
\cmidrule(lr){2-6} \cmidrule(lr){7-11}
 & Layer 4 & Layer 5 & Layer 6 & Layer 7 & Layer 8
 & Layer 4 & Layer 5 & Layer 6 & Layer 7 & Layer 8 \\
\midrule
1k  & 170 & 177 & 181 & 182 & 191
    & 168 & 176 & 181 & 182 & 191 \\
2k  & 177 & 178 & 182 & 183 & 194
    & 173 & 177 & 181 & 183 & 193 \\
3k  & 183 & 180 & 183 & 184 & 196
    & 178 & 179 & 182 & 184 & 195 \\
4k  & 188 & 182 & 184 & 186 & 199
    & 183 & 180 & 183 & 184 & 197 \\
5k  & 193 & 183 & 185 & 187 & 202
    & 187 & 181 & 184 & 185 & 199 \\
6k  & 198 & 185 & 186 & 188 & 204
    & 192 & 182 & 184 & 186 & 201 \\
7k  & 202 & 186 & 187 & 189 & 207
    & 196 & 184 & 185 & 187 & 203 \\
8k  & 206 & 187 & 187 & 190 & 209
    & 200 & 185 & 186 & 188 & 205 \\
9k  & 209 & 188 & 188 & 191 & 212
    & 205 & 186 & 187 & 189 & 207 \\
10k & 212 & 189 & 189 & 192 & 214
    & 208 & 187 & 187 & 190 & 209 \\
\bottomrule
\end{tabular}
}
\end{table*}
\paragraph{Evaluation Metrics}
We evaluate editing performance using three standard metrics \citep{fangalphaedit,lirethinking}. 
\textbf{Success Rate} measures whether the edited model assigns the highest probability to the target output for the input prompt.
\textbf{Generality} measures whether the edited model produces the target output to paraphrased input prompts.
\textbf{Locality} measures whether the edited model preserves the original output for unrelated inputs. Details are provided in Appendix~\ref{app:baseline_hahr}. 

\paragraph{Main Results}
Figure~\ref{fig:performance_cf} presents the editing performance  on the \textsc{CounterFact} dataset. Results on \textsc{ZsRE} are provided in Appendix~\ref{app:results}. Our findings reveal a clear performance hierarchy. First, history-regularized variants—the current state of the art—significantly outperform their history-agnostic counterparts, sustaining reliable performance over thousands of sequential edits, whereas history-agnostic methods typically degrade after only a few hundred \citep{thedewikibigedit}. Second, and more notably, our proposed batch-equivalent sequential editing method achieves the best long-term performance, maintaining high performance across all three metrics even after tens of thousands of edits and establishing a consistent gap over both history-regularized and history-agnostic baselines. For a more concrete numerical view, Table~\ref{tab:qwen_perf}
reports the results of MEMIT-BE on Qwen3-8B at different editing
scales, complementing the trends shown in Figure~\ref{fig:performance_cf}. The results further demonstrate the strong editing performance
and robustness of batch-equivalent sequential editing as the
number of edits increases. In principle, we should also compare our method with batch editors to verify empirical equivalence beyond theoretical equivalence. We leave this comparison to Appendix~\ref{app:precision_impl}.

\begin{figure}[t]
  \centering
  \includegraphics[width=1\linewidth]{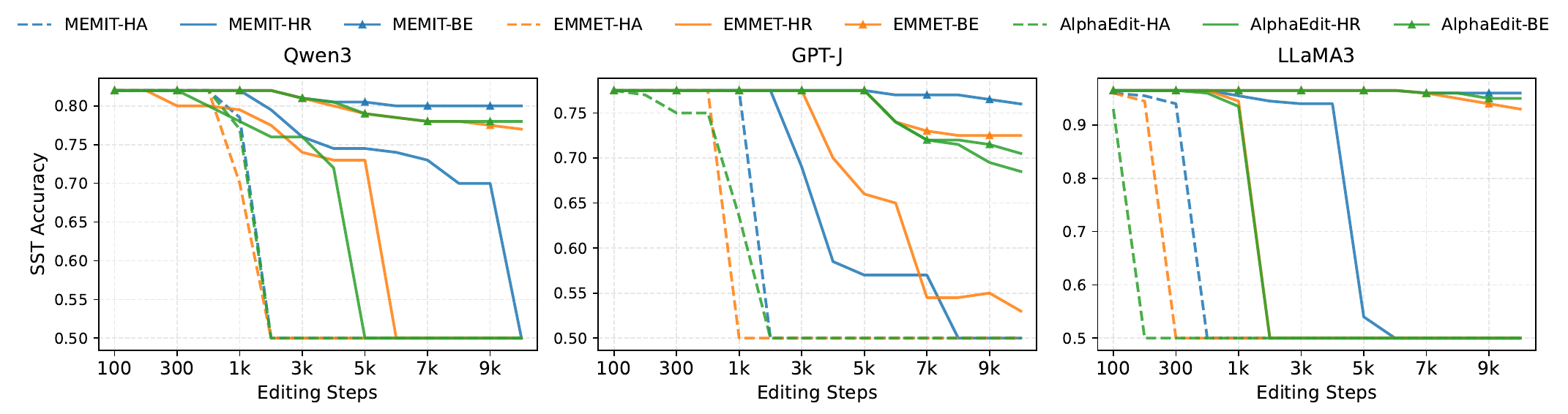}
\caption{
Evolution of language understanding capability (SST accuracy) during 10,000 sequential edits on the \textsc{CounterFact} dataset.
}
  \label{fig:sst_cf}
\end{figure}

\begin{figure}[t]
  \centering
  \includegraphics[width=1\linewidth]{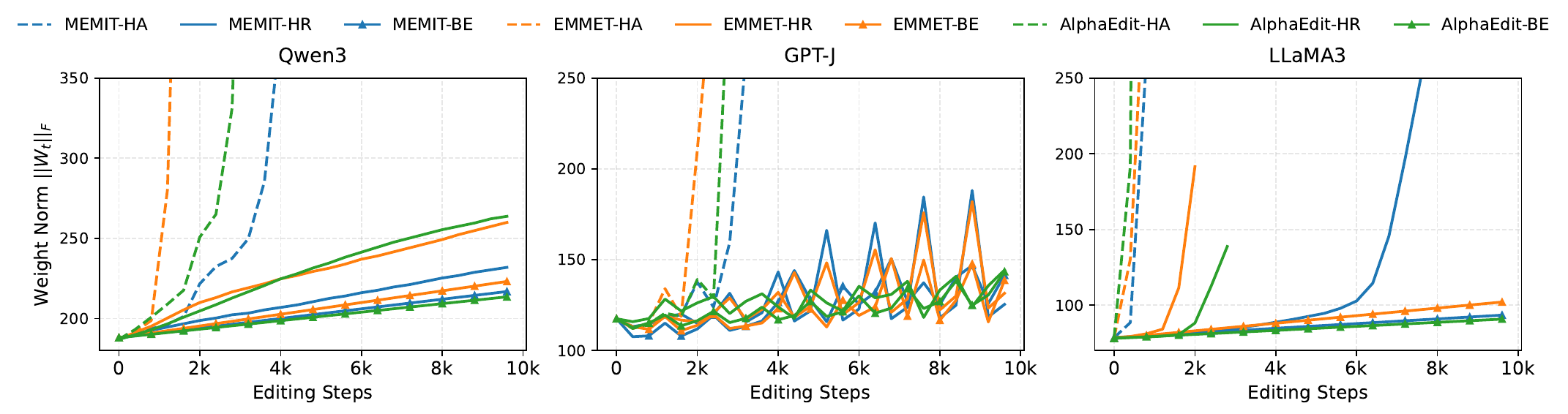}
\caption{Evolution of the weight norm during 10,000 sequential edits on the \textsc{CounterFact} dataset.}
  \label{fig:norm_cf}
\end{figure}

\subsection{Hierarchical Editing Stability}

Previous work has shown that editing methods suffer catastrophic performance degradation during long-term sequential editing, with success rates, generality, and locality all collapsing to zero \citep{gu2024model,gupta2024model,liu2025edit}. This raises a deeper question: does this failure reflect a decline limited to editing-specific metrics, or does it indicate a more fundamental breakdown in the model's overall behavior under sequential editing?

\paragraph{Evaluation Metrics}
To investigate this question, we monitor two complementary
aspects of model stability during sequential editing:
(i) the preservation of general-purpose language understanding,
evaluated using four benchmarks: SST-2,
MMLU, CoLA, and NLI, which assess sentiment classification,
general knowledge, linguistic acceptability, and natural
language inference, respectively \citep{fangalphaedit};
(ii) the evolution of the Frobenius norm
$\|\mathbf{W}_t\|_F$ of the edited weights at step $t$,
which characterizes the magnitude of parameter changes.
Additional stability metrics, including deviation from the
initial weights and leakage of pretrained knowledge, are
reported in Appendix~\ref{app:results}.

\paragraph{Main Results}
Figures~\ref{fig:sst_cf} and~\ref{fig:norm_cf} show SST-2 accuracy and weight norm evolution, respectively, on \textsc{CounterFact}.\footnote{Because the weight norm exhibits strong oscillations (especially for GPT-J), plotting every editing step yields dense, unreadable curves. We therefore subsample every 400 steps to reveal the underlying trend; full step-by-step results are provided in Appendix~\ref{app:results}.} Results on \textsc{ZSRE} are provided in Appendix~\ref{app:results}. We observe a strong correspondence between parameter drift and the degradation of general-purpose language understanding, consistent with the decline in editing-specific metrics reported earlier. As the edited weights increasingly deviate from their initial values, SST-2 accuracy tends to deteriorate. For example, with MEMIT-HR, the weight norm grows dramatically around 6,000 edits, coinciding with a sharp decline in both SST-2 accuracy and editing-specific performance.

\begin{wrapfigure}[19]{r}{0.43\textwidth}
    \centering
    \includegraphics[width=1.0\linewidth]{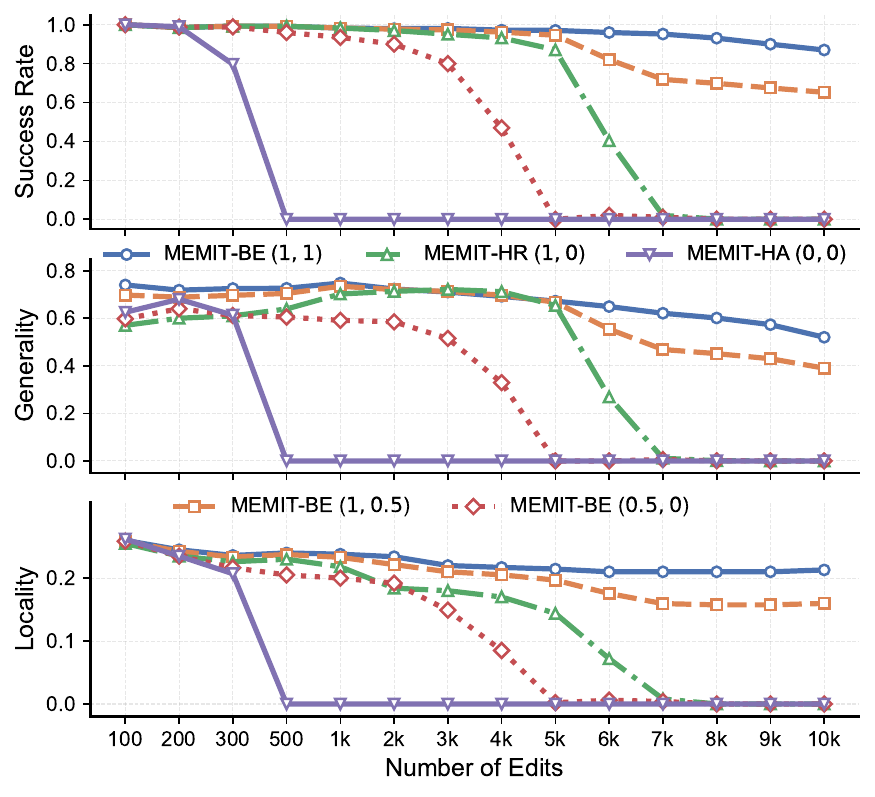}
\caption{Interpolation results for MEMIT-BE with different interpolation parameters $\beta_1$ and $\beta_2$ (values shown in parentheses).}
    \label{fig:memit_interpolation_llama_cf}
\end{wrapfigure}

Tables~\ref{tab:qwen_perf} and~\ref{tab:qwen_norm} provide
detailed numerical results for MEMIT-BE on Qwen3-8B across
different editing scales. The former evaluates the preservation
of general-purpose language understanding, while the latter
presents layer-wise weight norms across the five edited layers. These results show that batch-equivalent sequential editing preserves general-purpose language understanding while maintaining relatively stable layer-wise weight norms even after 10,000 edits.

Moreover, our results reveal a clear stability hierarchy among the three editing paradigms. History-agnostic editors exhibit rapid and unbounded weight growth, accompanied by a substantial degradation of language understanding within only a few hundred edits. History-regularized editors significantly mitigate parameter drift and better preserve model's general capabilities, but their performance could still deteriorate after thousands of edits, underperforming batch-equivalent sequential editors. These empirical trends are consistent with our theoretical
upper-bound analysis in Section~\ref{sec:sta}: history-agnostic
methods could exhibit linear weight perturbation growth, history-regularized methods admit a
logarithmic growth bound, and our method guarantees a
constant upper bound on weight perturbations. Note that the explosive weight growth observed in the \textit{near-collapse} phase does not contradict these bounds, as the theoretical upper bounds $C_k$ and $C_r$ on the key and residual vectors can scale with the model weights $\mathbf{W}_t$.

\begin{wrapfigure}[18]{r}{0.45\textwidth}
    \centering
    \includegraphics[width=1\linewidth]{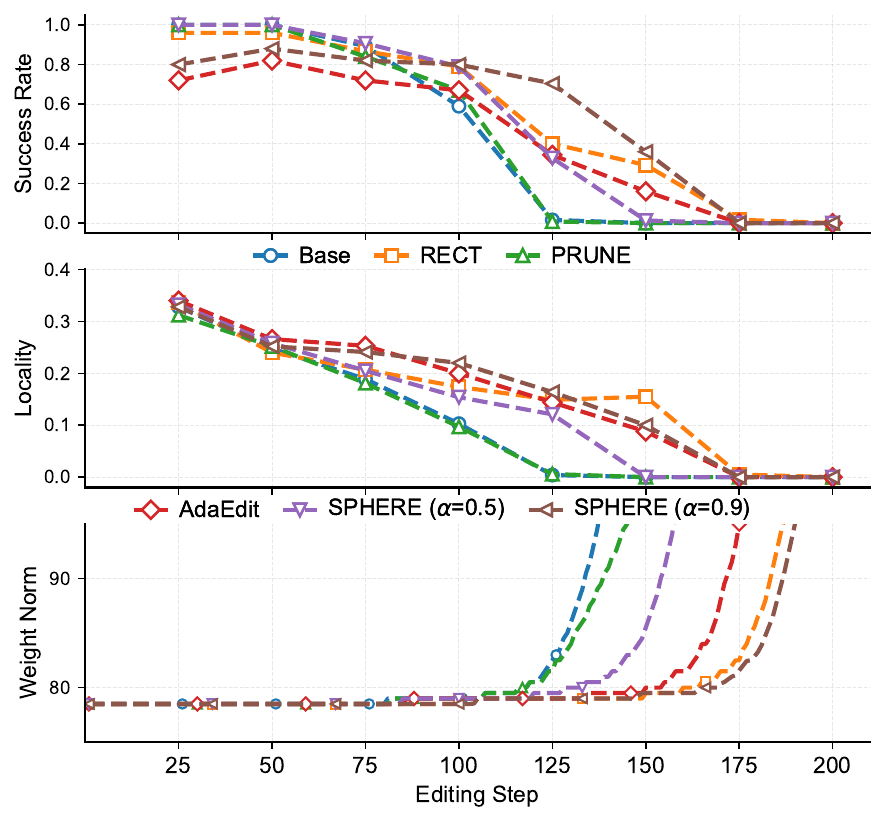}
    \caption{Effect of post-hoc regularization on AlphaEdit-HA (base), with $\alpha$ denoting the regularization strength.}
    \label{fig:post_reg_alphaedit_ha_llama_cf}
\end{wrapfigure}

\subsection{Interpolation Between Hierarchies}

We note that the hierarchical gap between history-agnostic,
history-regularized, and batch-equivalent methods is not
insurmountable. Instead, these methods can be unified
within a single formulation governed by two interpolation
parameters, $\beta_1$ and $\beta_2$, which control history
protection and batch-equivalence, respectively (see
Appendix~\ref{subapp:uni_bese} for details).  By varying these parameters,
the formulation accommodates a continuum of editing
behaviors. For example, setting $\beta_1=1$ and $\beta_2=0.5$ yields an intermediate
variant between batch-equivalent and history-regularized
editing. Figure~\ref{fig:memit_interpolation_llama_cf}
illustrates this interpolation for MEMIT-BE on the
\textsc{CounterFact} dataset using LLaMA. Results on \textsc{ZsRE} are provided in
Appendix~\ref{app:results}.

\subsection{Role of Post-Hoc Regularized Editing}
Theorem~\ref{thm:norm_growth} shows that
post-hoc regularization inherits the stability behavior
of the underlying sequential editor. We empirically examine
this effect in Figure~\ref{fig:post_reg_alphaedit_ha_llama_cf},
which compares four post-hoc regularization strategies
(\textsc{RECT}, \textsc{PRUNE}, \textsc{AdaEdit}, and
\textsc{SPHERE}) applied to AlphaEdit-HA on the
\textsc{CounterFact} dataset using LLaMA. These strategies
interpolate between the original update ($\alpha=0$) and
a zero update ($\alpha=1$), trading editing efficacy for
reduced weight perturbation. As shown in the figure, although post-hoc regularization
extends the editing horizon and slows down model collapse,
its benefits remain limited to a few
hundred edits. This leaves a substantial gap relative to
history-regularized sequential editors, which sustain stable editing performance across thousands of
updates.
Additional results and details on $\alpha$ are provided
in Appendix~\ref{app:results}.

\section{Conclusion and Limitations}

We have addressed a fundamental limitation in sequential knowledge editing within the \textit{locate-then-edit} (LTE) framework: the substantial performance gap between practical sequential updates and idealized batch editing. By designing a sequential algorithm that is exactly equivalent to batch editing, we close this gap, enabling tens of thousands of sequential edits. Furthermore, our convergence analysis reveals that long-term stability critically depends on controlling weight perturbations. This insight not only explains empirical observations of model degradation in prior methods but also provides a principled criterion for designing future sequential editing algorithms. Importantly, since batch LTE is secretly sequential LTE under specific editing objectives, our findings suggest that research efforts can productively refocus on advancing batch editing techniques—knowing that improvements there will directly carry over to the sequential setting. We outline several promising future research directions in Appendix~\ref{app:futurework}.

Our batch-equivalence guarantee requires the key-residual
pairs defining the editing objective to remain fixed
throughout the editing process, which can be achieved by
retaining the original model weights when extracting them.
However, this only guarantees batch equivalence under
single-layer editing or multi-layer editing with independent
editing objectives across layers. Similar settings are also
adopted in prior work on batch and sequential editing
equivalence~\citep{guotowards,wang2026the}. For multi-layer
editing algorithms where updates to earlier layers affect
the key-residual pairs of subsequent layers~\citep{mengmass,lirethinking,liubackward},
applying our method may deviate from the batch solution.
Nevertheless, our method retains strong performance and
can even empirically outperform batch editing. Appendix \ref{app:multi} evaluates our method on existing multi-layer editing algorithms with cross-layer dependencies against their batch counterparts.

\section*{Acknowledgments}
This work was supported by the National Key Research and Development Program of China under Grant 2024YFC3307900; the National Natural Science Foundation of China under Grants 62376103, 62302184, 62436003, 62472332, and 62672205; the Major Science and Technology Project of Hubei Province under Grants 2025BAB011, 2024BAA008, and 2024BAB039; the Hubei Science and Technology Talent Service Project under Grant 2024DJC078; the Science and Technology Innovation Talent Program under grant 2023CB005; and Wuhan Dameng Database Co., Ltd.

\bibliographystyle{plainnat}
\bibliography{ref}

\appendix

\section{Unified Batch Editor}
\label{app:unify}
Given the batch editor
\begin{equation}
\Delta_j^* = \lambda_1 \mathbf{R} \mathbf{K}^\top \left( \lambda_1 \mathbf{K} \mathbf{K}^\top + \lambda_2 \mathbf{K}_0 \mathbf{K}_0^\top \right)^{-1},
\label{eq:opt_editor_correct}
\end{equation}

We analyze the limiting case  when $\lambda_1\to\infty$ and $\lambda_2\to\infty$. 
\paragraph{Limiting Case 1: $\lambda_1\to\infty$}
Factor out  $ \lambda_1 $  from the inverse:
$$
\Delta_j^* 
= \mathbf{R} \mathbf{K}^\top \left( \mathbf{K} \mathbf{K}^\top + \frac{\lambda_2}{\lambda_1} \mathbf{M} \right)^{-1}.
$$
Now apply the matrix identity (a special case of the Woodbury formula):
$$
\mathbf{K}^\top \left( \mathbf{K} \mathbf{K}^\top + \epsilon \mathbf{M} \right)^{-1}
= \left( \mathbf{K}^\top \mathbf{M}^{-1} \mathbf{K} + \epsilon \mathbf{I} \right)^{-1} \mathbf{K}^\top \mathbf{M}^{-1}, \quad \epsilon > 0.
$$
Setting  $ \epsilon = \lambda_2 / \lambda_1 $ , we obtain
$$
\Delta_j^* = \mathbf{R} \left( \mathbf{K}^\top \mathbf{M}^{-1} \mathbf{K} + \frac{\lambda_2}{\lambda_1} \mathbf{I} \right)^{-1} \mathbf{K}^\top \mathbf{M}^{-1}.
$$
Taking the limit  $ \lambda_1 \to \infty $  (with  $ \lambda_2 > 0 $  fixed) gives  $ \frac{\lambda_2}{\lambda_1} \to 0 $. Hence,
$$
\lim_{\lambda_1 \to \infty} \Delta_j^*
= \mathbf{R} \left( \mathbf{K}^\top (\mathbf{K}_0 \mathbf{K}_0^\top)^{-1} \mathbf{K} \right)^{-1} \mathbf{K}^\top (\mathbf{K}_0 \mathbf{K}_0^\top)^{-1},
$$
which is exactly the update rule of ROME \citep{meng2022locating} and EMMET \citep{gupta2024unified}.
\paragraph{Limiting Case 2:  $ \lambda_2 \to \infty $}

To analyze the limiting behavior of the unified batch editor as  $\lambda_2 \to \infty$, we first introduce the following matrix-analytic result \citep{ben2003generalized}:

\begin{lemma}
\label{lem:null_space_limit}
Let  $ \mathbf{A} \in \mathbb{R}^{d \times d} $  be symmetric positive semidefinite, and let  $ \mathbf{K}_0 \in \mathbb{R}^{d \times m_0} $  be full column rank. Define the orthogonal projector onto the null space of  $ \mathbf{K}_0^\top $  as
$$
\mathbf{P} := \mathbf{I} - \mathbf{K}_0 (\mathbf{K}_0^\top \mathbf{K}_0)^{-1} \mathbf{K}_0^\top.
$$
Then
$$
\lim_{\lambda_2 \to \infty} \left( \mathbf{A} + \lambda_2 \, \mathbf{K}_0 \mathbf{K}_0^\top \right)^{-1}
= \mathbf{P} \left( \mathbf{P} \mathbf{A} \mathbf{P} \right)^{\dagger} \mathbf{P},
$$
where  $ {\dagger} $  denotes the Moore–Penrose pseudoinverse.
\end{lemma}

In our setting, set  $ \mathbf{A} = \lambda_1 \mathbf{K} \mathbf{K}^\top $ , which is symmetric positive semidefinite. Applying Lemma \ref{lem:null_space_limit}, we obtain
\begin{equation}
\lim_{\lambda_2 \to \infty} \left( \lambda_1 \mathbf{K} \mathbf{K}^\top + \lambda_2 \mathbf{K}_0 \mathbf{K}_0^\top \right)^{-1}
= \frac{1}{\lambda_1} \, \mathbf{P} \left( \mathbf{P} \mathbf{K} \mathbf{K}^\top \mathbf{P} \right)^{\dagger} \mathbf{P}.
\label{eq:inverse_limit}
\end{equation}

Substituting Eq. \eqref{eq:inverse_limit} into the expression for  $ \Delta_j^* $  in Eq. \eqref{eq:opt_editor_correct} yields
\begin{align}
\lim_{\lambda_2 \to \infty} \Delta_j^*
&= \lambda_1 \mathbf{R} \mathbf{K}^\top \cdot \frac{1}{\lambda_1} \, \mathbf{P} \left( \mathbf{P} \mathbf{K} \mathbf{K}^\top \mathbf{P} \right)^{\dagger} \mathbf{P} \nonumber \\
&= \mathbf{R} \mathbf{K}^\top \mathbf{P} \left( \mathbf{P} \mathbf{K} \mathbf{K}^\top \mathbf{P} \right)^{\dagger} \mathbf{P}\\
&= \mathbf{R} \mathbf{K}^\top \mathbf{P} \left( \mathbf{P} \mathbf{K} \mathbf{K}^\top \mathbf{P} \right)^{\dagger}.
\label{eq:with_trailing_P}
\end{align}
where the last equality follows because the range of  $ \left( \mathbf{P} \mathbf{K} \mathbf{K}^\top \mathbf{P} \right)^{\dagger} $  is contained in  $ \operatorname{col}(\mathbf{P}) $ , and  $ \mathbf{P} $  acts as the identity on its own column space.

In practice, to ensure numerical stability and invertibility, a small regularization term  $ \mu \mathbf{I} $  ( $ \mu > 0 $ ) is added, yielding the AlphaEdit-like update rule \citep{fangalphaedit}:
\begin{align}
\lim_{\lambda_2 \to \infty} \Delta_j^*
= \mathbf{R} \mathbf{K}^\top \mathbf{P} \left( \mathbf{P} \mathbf{K} \mathbf{K}^\top \mathbf{P}+ \mu \mathbf{I}\right)^{-1}.
\label{eq:with_trailing_P}
\end{align}
\section{Derivation of Batch-Equivalent Sequential Editing}
\label{app:bese}
It has long been believed in the model editing community that sequential \textit{Locate-then-Edit} (LTE) is inferior to batch LTE. However, we show below that they are in fact exactly equivalent—no assumptions are introduced in establishing the equivalence between sequential LTE and batch LTE.
\subsection{Derivation of Batch-Equivalent Sequential Editing for MEMIT (Theorem \ref{thm:batch_equivalent_memit})}
\label{subapp:memit}

We begin with the batch solution for $j$ edits:
\[
\Delta_j = B_j A_j^{-1},
\]
where
\[
A_j = \lambda_1 \sum_{i=1}^{j} k_i k_i^\top + \lambda_2 K_0 K_0^\top, \qquad 
B_j = \lambda_1 \sum_{i=1}^{j} r_i k_i^\top.
\]
Note that $B_j = \Delta_j A_j$, a relation we will use later.

\noindent\textbf{Step 1: Initial equivalence.} 
At the first step $j=1$, the sequential solution $\Delta_1'$ is computed directly from the first edit $(k_1, r_1)$:
\[
\Delta_1' = \lambda_1 r_1 k_1^\top (\lambda_1 k_1 k_1^\top + \lambda_2 K_0 K_0^\top)^{-1}
\]
which is trivially the batch solution $\Delta_1 = B_1 A_1^{-1}$.

\noindent\textbf{Step 2: Adding the $(j+1)$-th edit.}
For the $(j+1)$-th edit, the matrices update as:
\[
A_{j+1} = A_j + \lambda_1 k_{j+1} k_{j+1}^\top, \qquad 
B_{j+1} = B_j + \lambda_1 r_{j+1} k_{j+1}^\top.
\]
The batch solution after $j+1$ edits is therefore:
\[
\Delta_{j+1} = B_{j+1} A_{j+1}^{-1} = \left(B_j + \lambda_1 r_{j+1} k_{j+1}^\top\right) A_{j+1}^{-1}.
\]
Substituting $B_j = \Delta_j A_j$, we obtain:
\begin{equation}
\label{eq:subapp_recurse_memit}
\Delta_{j+1} = \left(\Delta_j A_j + \lambda_1 r_{j+1} k_{j+1}^\top\right) A_{j+1}^{-1}.
\end{equation}

\noindent\textbf{Step 3: Applying the Sherman–Morrison–Woodbury formula.}
The SMW formula states:
\[
(A_j + \lambda_1 k_{j+1} k_{j+1}^\top)^{-1} = A_j^{-1} - \frac{\lambda_1 A_j^{-1} k_{j+1} k_{j+1}^\top A_j^{-1}}{1 + \lambda_1 k_{j+1}^\top A_j^{-1} k_{j+1}}.
\]
Applying this to Eq.\ref{eq:subapp_recurse_memit} and simplifying:
\begin{equation}
\label{eq:subapp_recurse_memit1}
\begin{aligned}
\Delta_{j+1} &= \left(\Delta_j A_j + \lambda_1 r_{j+1} k_{j+1}^\top\right) \left( A_j^{-1} - \frac{\lambda_1 A_j^{-1} k_{j+1} k_{j+1}^\top A_j^{-1}}{1 + \lambda_1 k_{j+1}^\top A_j^{-1} k_{j+1}} \right) \\
&= \Delta_j + \lambda_1 (r_{j+1} - \Delta_j k_{j+1}) \underbrace{\frac{k_{j+1}^\top A_j^{-1}}{1 + \lambda_1 k_{j+1}^\top A_j^{-1} k_{j+1}}}_{\text{Term } X}.
\end{aligned}
\end{equation}

\noindent\textbf{Step 4: Simplifying Term $X$.}
Observe that Term $X$ is actually equal to $k_{j+1}^\top A_{j+1}^{-1}$:
\[
k_{j+1}^\top A_{j+1}^{-1} = k_{j+1}^\top \left( A_j^{-1} - \frac{\lambda_1 A_j^{-1} k_{j+1} k_{j+1}^\top A_j^{-1}}{1 + \lambda_1 k_{j+1}^\top A_j^{-1} k_{j+1}} \right)
\]
\[
= k_{j+1}^\top A_j^{-1} - \frac{\lambda_1 (k_{j+1}^\top A_j^{-1} k_{j+1}) k_{j+1}^\top A_j^{-1}}{1 + \lambda_1 k_{j+1}^\top A_j^{-1} k_{j+1}}
\]
\[
= \frac{k_{j+1}^\top A_j^{-1}}{1 + \lambda_1 k_{j+1}^\top A_j^{-1} k_{j+1}}.
\]
Thus Term $X = k_{j+1}^\top A_{j+1}^{-1}$.

\noindent\textbf{Step 5: The sequential update rule.}
Substituting this identity into Eq. \ref{eq:subapp_recurse_memit1}, we obtain:
\begin{equation}
\label{eq:subapp_recurse_memit_delta}
\Delta_{j+1} = \Delta_j + \lambda_1 (r_{j+1} - \Delta_j k_{j+1}) k_{j+1}^\top A_{j+1}^{-1}.
\end{equation}
Recalling $\mathbf{W}_{j+1} = \mathbf{W}_0 + \Delta_{j+1}$ and $\mathbf{W}_j = \mathbf{W}_0 + \Delta_j$, we convert to weight-space update:
\[
\mathbf{W}_{j+1}^* = \mathbf{W}_j^* + \lambda_1 \left(r_{j+1} - (\mathbf{W}_j^* - \mathbf{W}_0) k_{j+1}\right) k_{j+1}^\top A_{j+1}^{-1}.
\]
Rearranging yields the final form:
\begin{equation}
\label{eq:subapp_recurse_memit2}
\mathbf{W}_{j+1}^* = \mathbf{W}_j^* \left( I - \lambda_1 k_{j+1} k_{j+1}^\top A_{j+1}^{-1} \right) + \lambda_1 \left( r_{j+1} + \mathbf{W}_0 k_{j+1} \right) k_{j+1}^\top A_{j+1}^{-1}.
\end{equation}

\noindent This update rule defines the sequential operator $f^*$. Starting from $j=1$, it recursively computes $\mathbf{W}_{j+1}^*$ from $\mathbf{W}_{j}^*$ while exactly matching the batch solution at every step.

\subsection{Derivation of Batch-Equivalent Sequential Editing for AlphaEdit}
\label{subapp:alphaedit_bese}
We begin with the batch solution for $j$ edits:
\[
\Delta_j = B_j A_j^{-1},
\]
where
\[
A_j = \left( \sum_{i=1}^{j} k_i k_i^\top \right) P + \mu I, \qquad 
B_j = \left( \sum_{i=1}^{j} r_i k_i^\top \right) P.
\]
Note that $B_j = \Delta_j A_j$, a relation we will use later.

\noindent\textbf{Step 1: Initial equivalence.} 
At the first step $j=1$, the sequential solution $\Delta_1'$ is computed directly from the first edit $(k_1, r_1)$:
\[
\Delta_1' = r_1 k_1^\top P (k_1 k_1^\top P + \mu I)^{-1}
\]
which is trivially the batch solution $\Delta_1 = B_1 A_1^{-1}$.

\noindent\textbf{Step 2: Adding the $(j+1)$-th edit.}
For the $(j+1)$-th edit, the matrices update as:
\[
A_{j+1} = A_j + k_{j+1} k_{j+1}^\top P, \qquad 
B_{j+1} = B_j + r_{j+1} k_{j+1}^\top P.
\]
The batch solution after $j+1$ edits is therefore:
\[
\Delta_{j+1} = B_{j+1} A_{j+1}^{-1} = \left(B_j + r_{j+1} k_{j+1}^\top P\right) A_{j+1}^{-1}.
\]
Substituting $B_j = \Delta_j A_j$, we obtain:
\begin{equation}
\label{eq:alpharecurse}
\Delta_{j+1} = \left(\Delta_j A_j + r_{j+1} k_{j+1}^\top P\right) A_{j+1}^{-1}.
\end{equation}

\noindent\textbf{Step 3: Applying the Sherman–Morrison–Woodbury formula.}
The SMW formula states:
\[
(A_j + k_{j+1} k_{j+1}^\top P)^{-1} = A_j^{-1} - \frac{A_j^{-1} k_{j+1} k_{j+1}^\top P A_j^{-1}}{1 + k_{j+1}^\top P A_j^{-1} k_{j+1}}.
\]
Applying this to Eq.\ref{eq:alpharecurse} and simplifying:
\begin{equation}
\label{eq:alpharecurse1}
\begin{aligned}
\Delta_{j+1} &= \left(\Delta_j A_j + r_{j+1} k_{j+1}^\top P\right) \left( A_j^{-1} - \frac{A_j^{-1} k_{j+1} k_{j+1}^\top P A_j^{-1}}{1 + k_{j+1}^\top P A_j^{-1} k_{j+1}} \right) \\
&= \Delta_j + (r_{j+1} - \Delta_j k_{j+1}) \underbrace{\frac{k_{j+1}^\top P A_j^{-1}}{1 + k_{j+1}^\top P A_j^{-1} k_{j+1}}}_{\text{Term } X}.
\end{aligned}
\end{equation}

\noindent\textbf{Step 4: Simplifying Term $X$.}
Observe that Term $X$ is actually equal to $k_{j+1}^\top P A_{j+1}^{-1}$:
\[
\begin{aligned}
k_{j+1}^\top P A_{j+1}^{-1} &= k_{j+1}^\top P \left( A_j^{-1} - \frac{A_j^{-1} k_{j+1} k_{j+1}^\top P A_j^{-1}}{1 + k_{j+1}^\top P A_j^{-1} k_{j+1}} \right) \\
&= k_{j+1}^\top P A_j^{-1} - \frac{(k_{j+1}^\top P A_j^{-1} k_{j+1}) k_{j+1}^\top P A_j^{-1}}{1 + k_{j+1}^\top P A_j^{-1} k_{j+1}} \\
&= \frac{k_{j+1}^\top P A_j^{-1}}{1 + k_{j+1}^\top P A_j^{-1} k_{j+1}}.
\end{aligned}
\]
Thus Term $X = k_{j+1}^\top P A_{j+1}^{-1}$.

\noindent\textbf{Step 5: The sequential update rule.}
Substituting this identity into Eq. \ref{eq:alpharecurse1}, we obtain:
\[
\Delta_{j+1} = \Delta_j + (r_{j+1} - \Delta_j k_{j+1}) \, k_{j+1}^\top P A_{j+1}^{-1}.
\]

\noindent Recalling $\mathbf{W}_j = \mathbf{W}_0 + \Delta_j$ and $\mathbf{W}_{j+1} = \mathbf{W}_0 + \Delta_{j+1}$, we convert to weight-space update:
\[
\mathbf{W}_{j+1} = \mathbf{W}_j + (r_{j+1} - (\mathbf{W}_j - \mathbf{W}_0) k_{j+1}) \, k_{j+1}^\top P A_{j+1}^{-1}.
\]

\noindent Rearranging yields the final form:
\[
\mathbf{W}_{j+1} = \mathbf{W}_j \left( I - k_{j+1} k_{j+1}^\top P A_{j+1}^{-1} \right) + (r_{j+1} + \mathbf{W}_0 k_{j+1}) \, k_{j+1}^\top P A_{j+1}^{-1}.
\]

\noindent This update rule defines the sequential operator for AlphaEdit. Starting from $j=1$, it recursively computes $\mathbf{W}_{j+1}$ from $\mathbf{W}_j$ while exactly matching the batch solution at every step.
\begin{algorithm}[t]
\caption{Batch-Equivalent Sequential Editing (MEMIT)}
\label{alg:seq_edit}
\begin{algorithmic}[1]
\REQUIRE Initial weight $\mathbf{W}_0$ and pretrained knowledge key covariance matrix $\mathbf{C}_0 = \mathbf{K}_0 \mathbf{K}_0^\top$.
\ENSURE Updated weights  $ \mathbf{W}_j $  after each edit
\STATE Compute  $ A_0^{-1} \gets (\lambda_2\mathbf{C}_0)^{-1} $
\STATE  $ \Delta_0 \gets \mathbf{0} $ 
\FOR{ $ j = 1 $  to  $ t $ }
    \STATE Recieve new fact $(s_i,r_i,o_i)$
    \STATE Compute key-value pairs $ (k_j, r_j) $ based on $\mathbf{W}_0$ 
    \STATE Update inverse via SMW:
$$
    A_j^{-1} \gets A_{j-1}^{-1} - \frac{A_{j-1}^{-1} k_j k_j^\top A_{j-1}^{-1}}{1/\lambda_1 + k_j^\top A_{j-1}^{-1} k_j}
$$
    \STATE Compute update matrix:
$$
    \Delta_j \gets \Delta_{j-1} + \lambda_1 (r_j - \Delta_{j-1} k_j) k_j^\top A_j^{-1}
$$
    \STATE Set  $ \mathbf{W}_j \gets \mathbf{W}_0 + \Delta_j $ 
\ENDFOR
\end{algorithmic}
\end{algorithm}

\subsection{Unified Sequential Editor}
\label{subapp:uni_bese}
We demonstrate that history-agnostic (Section \ref{sec:sta}), history-regularized (Section \ref{sec:sta}), and batch-equivalent sequential editing (Section \ref{sec:method}) can be unified into a single formulation. Without loss of generality, we take MEMIT as an example. Specifically, to generalize the sequential update (Eq.~\ref{eq:subapp_recurse_memit2}), we augment it with two interpolation parameters: $\beta_1$ controls the protection of editing history, while $\beta_2$ controls batch-equivalence.
\begin{equation}
\label{eq:subapp_recurse_memit_ge}
\mathbf{W}_{j+1}^* =  \mathbf{W}_j^* \left( I - \beta_2\lambda_1 k_{j+1} k_{j+1}^\top A_{j+1}^{-1} \right) + \lambda_1 \left( r_{j+1} + \beta_2 \mathbf{W}_0 k_{j+1} \right) k_{j+1}^\top A_{j+1}^{-1},
\end{equation}
where
\[
A_j = k_j k_j^\top+ \beta_1 \sum_{i=1}^{j-1} k_i k_i^\top + \lambda_2 \mathbf{K}_0 \mathbf{K}_0^\top, \qquad 
B_j = \sum_{i=1}^{j} r_i k_i^\top.
\]
Setting $\beta_1 = \beta_2 = 1$ recovers the batch-equivalent sequential editor in Eq.~\ref{eq:subapp_recurse_memit2}. More generally, the unified formulation can degenerate to existing editors by varying $\beta_1, \beta_2 \in [0,1]$. For simplicity, we follow MEMIT's default setting of $\lambda_1 = 1$ and rewrite Eq.~\ref{eq:subapp_recurse_memit_ge} from the perspective of weight perturbation (Eq.~\ref{eq:subapp_recurse_memit_delta}):

\begin{equation}
\label{eq:subapp_recurse_memit_ge_perturb}
\Delta_{j+1} = \Delta_j + (r_{j+1} - \beta_2\Delta_j k_{j+1}) k_{j+1}^\top A_{j+1}^{-1}.
\end{equation}

\textbf{History-Regularized Editor ($\beta_1 = 1,\beta_2=0$).}
Setting $\beta_2 = 0$ eliminates the third term in the expanded form of the update rule (Eq. \ref{eq:subapp_recurse_memit_ge_perturb}), yielding:
\begin{equation}
\Delta_{j+1} = \Delta_j + r_{j+1} k_{j+1}^\top A_{j+1}^{-1},
\end{equation}
where $A_{j+1} = k_{j+1} k_{j+1}^\top+ \beta_1 \sum_{i=1}^{j} k_i k_i^\top + \lambda_2 \mathbf{K}_0 \mathbf{K}_0^\top$ remains unchanged, with $\beta_1>0$ (typically set to $1$ \citep{fangalphaedit,liubetaedit}) to preserve the full editing history.

\textbf{History-Agnostic Editor ($\beta_1 = 0,\beta_2 = 0$).}
If $A_{j+1}$ ignores previous edits (i.e., setting $\beta_1 = 0$ so that $A_{j+1} =  k_{j+1} k_{j+1}^\top + \lambda_2 \mathbf{K}_0 \mathbf{K}_0^\top$), the method reduces to repeatedly applying the vanilla MEMIT update in a sequential manner:
\begin{equation}
\Delta_{j+1} = \Delta_j + r_{j+1} k_{j+1}^\top \left( k_{j+1}k_{j+1}^\top + \lambda_2 \mathbf{K}_0 \mathbf{K}_0^\top \right)^{-1}.
\end{equation}
This approach is effectively \textbf{history-agnostic}, as it disregards prior edits when performing new ones \citep{gu2024model,thedewikibigedit}.

Based on the batch-equivalent sequential editor ($\beta_1=1,\beta_2=1$), we further have the following discussions.

\textbf{Batch MEMIT.}
Setting $j=0$ (so $\Delta_j = \mathbf{0}$) and replacing vectors with matrices recovers the \textbf{batch} MEMIT formulation:
\begin{equation}
\Delta = \mathbf{R} \mathbf{K}^\top \left( \mathbf{K} \mathbf{K}^\top +\lambda_2 \mathbf{K}_0 \mathbf{K}_0^\top \right)^{-1}.
\end{equation}

\textbf{Post-Processing Interpretation.}
If we approximate $r_{j+1}$ as $\bar\Delta_{j+1} k_{j+1}$, where $\bar\Delta_{j+1}$ is the update matrix from a third-party editor, the unified rule becomes:
\begin{equation}
\Delta_{j+1} = \Delta_j + \bar\Delta_{j+1} k_{j+1} k_{j+1}^\top A_{j+1}^{-1} - \Delta_j k_{j+1} k_{j+1}^\top A_{j+1}^{-1}.
\end{equation}
This can be viewed as a post-processing technique that transforms a vanilla update $\bar\Delta_{j+1}$ into an update that protects previous edits and pretrained knowledge. Such an idea is adopted in \textit{Simulating Ideal Editor} (SimIE \citep{guotowards}); however, due to approximation errors and propagation across multiple layers, SimIE empirically underperforms compared to history-regularized methods like AlphaEdit~\citep{fangalphaedit}. Moreover, in SimIE, $A_{j+1}$ is set to $\mathbf{K}_{[1:j+1]}\mathbf{K}_{[1:j+1]}^{\top} + \mathbf{I}$, which fails to preserve the pre-trained knowledge $\mathbf{K}_0$. Notably, the identity matrix $\mathbf{I}$ — rather than $\mathbf{K}_0\mathbf{K}_0^{\top}$ — is mandatory in the derivation of SimIE.

\subsection{Extension to Mini-Batch Sequential Editing}
\label{app:mini_batch}
Suppose that at step  $ j $ , a mini-batch of  $ b $  edits arrives simultaneously, represented by key and residual matrices  $ \mathbf{K}_j \in \mathbb{R}^{d_{\text{in}} \times b} $  and  $ \mathbf{R}_j \in \mathbb{R}^{d_{\text{out}} \times b}$, respectively.  
The Gram and cross-covariance matrices are then updated as
$$
A_{j+1} = A_j + \lambda_1 \mathbf{K}_j \mathbf{K}_j^\top, \qquad
B_{j+1} = B_j + \lambda_1 \mathbf{R}_j \mathbf{K}_j^\top,
$$
where  $ A_j = \lambda_1 \sum_{i=1}^{j-1} \mathbf{K}_i \mathbf{K}_i^\top + \lambda_2 \mathbf{K}_0 \mathbf{K}_0^\top $  and  $ B_j = \sum_{i=1}^{j-1} \lambda_1\mathbf{R}_i \mathbf{K}_i^\top $  follow the definitions in Eq.~\eqref{eq:solj}.
The inverse  $ A_{j+1}^{-1} $  can be computed efficiently using the Woodbury matrix identity:
$$
A_{j+1}^{-1} = A_j^{-1} - A_j^{-1} \mathbf{K}_j \left( \frac{1}{\lambda_1} \mathbf{I}_b + \mathbf{K}_j^\top A_j^{-1} \mathbf{K}_j \right)^{-1} \mathbf{K}_j^\top A_j^{-1},
$$
which reduces to the Sherman–Morrison formula when  $ b = 1 $.  
The update for  $ \Delta_{j+1} $  remains structurally identical to the single-edit case:
$$
\Delta_{j+1} = \Delta_j + (\mathbf{R}_j - \Delta_j \mathbf{K}_j) \left( \frac{1}{\lambda_1} \mathbf{I}_b + \mathbf{K}_j^\top A_j^{-1} \mathbf{K}_j \right)^{-1} \mathbf{K}_j^\top A_j^{-1}.
$$
Thus, the core algorithm extends naturally to mini-batch sequential editing with only a modest increase in per-step computation (dominated by solving a  $ b \times b $  linear system). When  $ b \ll d_{\text{in}} $, this remains highly efficient; when $b = d_{\text{in}}$, it incurs the full $O(d_{\text{in}}^3)$ matrix inversion cost, recovering the naive  implementation cost without SMW. 

\begin{table*}[!t]
\centering
\caption{Comparison of three sequential implementation
strategies against the batch editor on Qwen3-8B
under \texttt{float32} precision. Editing performance is
reported after 2,000 and 10,000 sequential edits on
\textsc{CounterFact} and \textsc{ZSRE}.}
\label{tab:results_precision}
\setlength{\tabcolsep}{4pt}
\resizebox{0.95\textwidth}{!}{
\begin{tabular}{l c c c c c c c c c c c c}
\toprule
\multirow{2}{*}{\textbf{Method}} &
\multicolumn{6}{c}{\textbf{CounterFact}} &
\multicolumn{6}{c}{\textbf{ZSRE}} \\
\cmidrule(lr){2-7} \cmidrule(lr){8-13}
& \multicolumn{3}{c}{\textbf{$T=2000$}}
& \multicolumn{3}{c}{\textbf{$T=10,000$}}
& \multicolumn{3}{c}{\textbf{$T=2000$}}
& \multicolumn{3}{c}{\textbf{$T=10,000$}} \\
\cmidrule(lr){2-4} \cmidrule(lr){5-7}
\cmidrule(lr){8-10} \cmidrule(lr){11-13}
& \textbf{Suc.} & \textbf{Gen.} & \textbf{Loc.}
& \textbf{Suc.} & \textbf{Gen.} & \textbf{Loc.}
& \textbf{Suc.} & \textbf{Gen.} & \textbf{Loc.}
& \textbf{Suc.} & \textbf{Gen.} & \textbf{Loc.} \\
\midrule
Batch
& 97.6& 53.1& 10.8
& 91.3& 49.0 &8.8
& 94.5& 88.0& 38.1
& 90.9& 84.5 &37.2 \\

SMW 
& 95.3& 52.3& 10.2
& 81.2& 44.5 &7.2
& 91.5& 87.0& 38.6
& 83.9& 79.5 &37.1 \\

Direct Inverse 
& 96.9& 52.8& 10.6
& 88.2& 46.7 &8.1
& 94.1& 87.5& 38.4
& 88.1& 82.7 &37.7 \\

Direct Batch-Form
& 97.6& 53.1& 10.8
& 91.3& 49.0 &8.8
& 94.5& 88.0& 38.1
& 90.9& 84.5 &37.2 \\
\bottomrule
\end{tabular}}
\end{table*}
\begin{table}[t!]
\centering
\caption{Comparison of editing time for 10,000 sequential edits on LLaMA-3-8B.}
\begin{tabular}{l|c|c|c|c|c|c}
\hline
Method & MEMIT & EMMET& AlphaEdit & RECT & PRUNE & Ours \\
\hline
Time (hours) & 16 & 15 & 18 & 16 & 28 & 9 \\
\hline
\end{tabular}
\label{tab:time_comparison}
\end{table}
\section{Practical Implementation of Batch Equivalent Sequential Editing}
\label{app:precision_impl}

\subsection{Three Implementation Strategies}

Although Theorem~\ref{thm:batch_equivalent_memit} establishes
the equivalence between sequential and batch editing under
exact arithmetic, different implementations of the sequential
operator may exhibit different numerical behaviors in
finite-precision arithmetic. In particular, the recursive
computation of the inverse matrix and the weight matrix may
introduce numerical errors that propagate across editing
steps. To investigate this issue, we consider three
implementation strategies, all based on the same theoretical
formulation. Without loss of generality, we illustrate using the batch-equivalent sequential operator derived from the MEMIT-style editing objective.
\paragraph{(1) SMW-based Recursive Implementation.}

We exploit the Sherman--Morrison--Woodbury (SMW) identity
to recursively update $A_{j+1}^{-1}$, and use
Theorem~\ref{thm:batch_equivalent_memit} to recursively
update the weight matrix $\mathbf{W}_{j+1}^*$. The resulting
algorithm is presented in Algorithm~\ref{alg:seq_edit}.

\paragraph{(2) Direct Inverse with Recursive Weight Updates.}

We directly recompute $A_{j+1}^{-1}$ at each editing step,
while still using Theorem~\ref{thm:batch_equivalent_memit}
to recursively update the weight matrix $\mathbf{W}_{j+1}^*$.

\paragraph{(3) Direct Batch-Form Implementation.}

We incrementally maintain $A_j$ and $B_j$ using
Eq.~\ref{eq:solj}, and directly compute the
weight matrix from the original weights at each step:
\begin{equation}
\mathbf{W}_{j+1}^*
=
\mathbf{W}_0+B_{j+1}A_{j+1}^{-1}.
\label{eq:direct_batch_form}
\end{equation}

The three implementations differ in how the inverse
matrix and weight matrix are computed: the first uses SMW
to recursively update the inverse and then recursively
updates the weights; the second recomputes the inverse
from scratch but still recursively updates the weights;
and the third computes the weight update from the
initial weight. Consequently,
the first two implementations may accumulate numerical
errors through recursive updates, whereas the third
avoids such error propagation. 
We compare their numerical deviations from batch editing
in Section \ref{app:ncbe}.

\subsection{Numerical Comparison with Batch Editing}
\label{app:ncbe}
We compare the three implementation strategies against the batch editor to investigate their numerical
behaviors under finite-precision arithmetic. Specifically, we evaluate their editing performance
using the standard \texttt{float32} precision for weight-matrix computations. Table ~\ref{tab:results_precision} reports the results
after 2k and 10k sequential edits on \textsc{CounterFact} and  \textsc{ZSRE}. The results show that the
numerical discrepancies increase with the number of sequential edits. In particular, the SMW-based
implementation exhibits the largest deviations, followed by the direct-inverse implementation. The
direct batch-form implementation matches the batch baseline. These findings illustrate that although
all three strategies implement the same batch-equivalent sequential formulation under exact arithmetic,
their practical numerical behaviors depend on how the recursive computations are carried out. 

We note that the batch-equivalence guarantee of our sequential algorithm, even for the direct batch-form implementation, applies only to single-layer editing or multi-layer editing with independent editing objectives across layers---the setting used in our main experiments. It does not directly extend to multi-layer editing with cross-layer dependencies, where updates to earlier layers alter the key-residual pairs of subsequent layers. We discuss this setting and its implications in Appendix~\ref{app:multi}.


\subsection{Computational Efficiency of SMW-based Implementation}

We analyze the practical efficiency of the
SMW-based implementation in terms of memory footprint
and computational time, and compare it with existing editing
methods.

\paragraph{Memory.}

As shown in Alg.~\ref{alg:seq_edit}, our SMW-based
implementation stores two matrices: $\mathbf{W}_0$ of size
$(d_{\text{in}},d_{\text{out}})$ and $A_j^{-1}$ of size
$(d_{\text{in}},d_{\text{in}})$. For LLaMA-3-8B
($d_{\text{in}}=14336$, $d_{\text{out}}=4096$) with
\texttt{float32} precision, these require approximately
0.2~GB and 0.8~GB, respectively. Baseline methods store
$\mathbf{C}_0=\mathbf{K}_0\mathbf{K}_0^\top$ of size
$(d_{\text{in}},d_{\text{in}})$, so our additional memory
cost is only the 0.2~GB required for $\mathbf{W}_0$.
Editing five layers (Section~\ref{sec:exp}) adds
$0.2\times5=1$~GB, which is negligible compared with the
32~GB required to load LLaMA-3-8B. Note that the memory
cost is independent of the number of edits.

\paragraph{Time.}

We exploit the SMW identity to efficiently maintain
$A_j^{-1}$ across editing steps, avoiding repeated
matrix inversions. We benchmark different editing methods
on LLaMA-3-8B using 10,000 sequential edits from
CounterFact (batch size = 1) on an NVIDIA A6000 48GB GPU.
As reported in Table~\ref{tab:time_comparison}, our method
completes all edits in approximately 9 hours, outperforming
the existing baselines. In contrast, baseline methods
require repeatedly inverting a $d_{\text{in}}\times
d_{\text{in}}$ matrix at each edited layer, taking
approximately 3 seconds per edit across five layers.
This amounts to 8--10 hours for 10,000 edits, in addition
to approximately 8 hours for extracting the $(k,v)$ pairs
using \texttt{bfloat16} precision. PRUNE incurs further
overhead from post-hoc regularization, resulting in a
total runtime exceeding 24 hours.

\section{Related Work on Stability Analysis}
\label{app:sta}

Four existing works have investigated the stability of model editing from different perspectives. PRUNE~\citep{maperturbation} examines the condition number of the weight matrix \(W\), arguing that constraining it can help mitigate the perturbation of previously stored knowledge. However, its analysis is static and focuses on individual edits, without characterizing how the weight matrix evolves over successive editing steps.

SPHERE~\citep{liu2026energyregularized} investigates the output perturbation of pretrained knowledge, quantified by \(\Delta W V\), where \(V\) denotes the pretrained knowledge matrix. It argues that preserving the uniformity of $W$ helps mitigate unintended changes to pretrained knowledge. Nevertheless, its analysis is also static and does not explicitly characterize the evolution of the weight matrix \(W_t\) throughout the sequential editing process.

BetaEdit~\citep{liubetaedit} explicitly considers the dynamics of cumulative weight perturbations, \(W_t-W_0\), during sequential editing. It compares history-agnostic and history-aware editing methods in terms of their accumulated perturbations. However, its analysis primarily focuses on comparing the relative perturbation magnitudes between these two classes, rather than providing a separate characterization of their general weight dynamics.

Similar to our work, HSE~\citep{fang2025hippocampal} studies the evolution of the weight matrix \(W_t\) through its Frobenius norm \(\|W_t\|_F\). However, its analysis relies on strong and idealized assumptions, such as mutual orthogonality among edit keys \(\{k_i\}_{i=1}^t\), which are generally difficult to satisfy in practical sequential editing scenarios. Moreover, its theoretical analysis is specific to HSE and does not characterize the stability of other sequential editing paradigms.

In contrast, our work provides a unified and systematic characterization of weight dynamics across multiple sequential editing paradigms, including history-agnostic, history-regularized, batch-equivalent, and post-hoc regularized editing. Without relying on rigid assumptions such as mutual orthogonality of edit keys, we establish distinct upper bounds on weight growth for different editing strategies, enabling a more comprehensive understanding and practical comparison of their long-term editing stability.

\section{Convergence Analysis for Sequential Editing Strategies}
\label{app:convergence}
We analyze the growth of the Frobenius norm of the weight matrix  $ W_t$ under three sequential update rules, assuming the structural conditions stated in the main text:
\begin{itemize}
    \item \textbf{Boundness:}  $\|k_t\|_2 \leq C_k $  and  $\|r_t\|_2 \leq C_r $  for all  $ t $ ;
    \item  \textbf{Invertibility:} The input covariance satisfies  $ \mathbb{E}[k k^\top] \succeq \alpha \mathbf{I} $  for some constant  $ \alpha > 0 $ . Equivalently, the empirical Gram matrix becomes well-conditioned asymptotically: $\frac{1}{t} K_{[1:t]} K_{[1:t]}^\top \succeq \alpha \mathbf{I}, \frac{1}{n} K_{0} K_{0}^\top \succeq \alpha \mathbf{I}\quad \text{as } t,n \to \infty,$ where  $K_{[1:t]}\in \mathbb{R}^{d_{in}\times t}$ denotes the key vector encoding the editing knowledge and $K_{0} \in \mathbb{R}^{d_{in}\times n}$ denotes the key vector encoding the pretrained knowledge.
\end{itemize}

\subsection{Upper Bounds}
\begin{proof}

\textbf{(1) History-Agnostic Sequential Editing.}  Define the per-step gain magnitude as
$$
g_t^{(1)} := \left\| A_t^{-1} k_t \right\|_2 = \left\| \left( k_t k_t^\top + K_0 K_0^\top \right)^{-1} k_t \right\|_2.
$$
We now derive the upper bound on  $g_t^{(1)}$. Let  $ P_0 := K_0 K_0^\top \succ 0 $ . By the Sherman--Morrison formula,
$$
(k_t k_t^\top + P_0)^{-1} k_t = \frac{1}{1 + k_t^\top P_0^{-1} k_t} P_0^{-1} k_t.
$$
Hence,
$$
g_t^{(1)} = \frac{\|P_0^{-1} k_t\|_2}{1 + k_t^\top P_0^{-1} k_t}.
$$
Let  $ z_t := P_0^{-1/2} k_t $. Then  $ \|P_0^{-1} k_t\|_2 = \|P_0^{-1/2} z_t\|_2 \le \lambda_{\min}(P_0)^{-1/2} \|z_t\|_2$, and  $ k_t^\top P_0^{-1} k_t = \|z_t\|_2^2$. ($\lambda_{\min}(P_0)$ is the smallest eigenvalues of $P_0$.) Therefore,
$$
g_t^{(1)} \le \frac{\lambda_{\min}(P_0)^{-1/2} \|z_t\|_2}{1 + \|z_t\|_2^2}.
$$
The scalar function  $ f(s) = s/(1+s^2) $  for  $ s \ge 0 $  attains its maximum value  $ 1/2 $  at  $ s=1$. Thus,
$$
g_t^{(1)} \le \frac{1}{2} \lambda_{\min}(P_0)^{-1/2}.
$$
By the triangle inequality, the Frobenius norm of  $ W_t $  satisfies the universal upper bound
$$
\|W_t\|_F \le \|W_0\|_F + \sum_{i=1}^t \left\| r_i k_i^\top A_i^{-1} \right\|_F
\le \|W_0\|_F + t \cdot \frac{C_r}{2} \lambda_{\min}(P_0)^{-1/2},
$$

\textbf{(2) History-Regularized Sequential Editing.} The update rule is
$$
W_t = W_{t-1} + r_t k_t^\top A_t^{-1},
$$
which unrolls to  $ W_t = W_0 + \sum_{s=1}^t r_s k_s^\top A_s^{-1} $. Hence,
\begin{align*}
\|W_t\|_F 
&\leq \|W_0\|_F + \sum_{s=1}^t \|r_s k_s^\top A_s^{-1}\|_F \\
&\leq \|W_0\|_F + \sum_{s=1}^t \|r_s\|_2 \|k_s\|_2 \|A_s^{-1}\|_2 \\
&\leq \|W_0\|_F + \sum_{s=1}^t \frac{C_r C_k}{s \alpha}
\leq \|W_0\|_F + \frac{C_r C_k}{\alpha}(1 + \log t).
\end{align*}
where $ \|A_s^{-1}\|_2 \leq \frac{1}{s \alpha}$ as we have
$$
\frac{1}{s} K_{[1:s]} K_{[1:s]}^\top \succeq \alpha I \quad \Rightarrow \quad A_s \succeq s \alpha I.
$$
\textbf{(3) Batch-Equivalent Sequential Editing.} The corresponding batch solution at every editing step is,
$$
W_t = W_0 + R_{[1:t]} K_{[1:t]}^\top A_t^{-1}.
$$
Using submultiplicativity of the Frobenius norm,
\begin{align*}
\|W_t\|_F 
&\leq \|W_0\|_F + \|R_{[1:t]}\|_F \cdot \|K_{[1:t]}^\top A_t^{-1}\|_2 \\
&\leq \|W_0\|_F + (\sqrt{t} C_r) \cdot \left( \|K_{[1:t]}\|_2 \cdot \|A_t^{-1}\|_2 \right) \\
&\leq \|W_0\|_F + \sqrt{t} C_r \cdot \left( \sqrt{t} C_k \cdot \frac{1}{t \alpha} \right)
= \|W_0\|_F + \frac{C_r C_k}{\alpha}.
\end{align*}
\textbf{(4) Post-Hoc Regularized Editing.}
We assume that the post-hoc regularization operator is non-expansive with respect to the Frobenius norm, i.e., \(\|\operatorname{PR}(\Delta,\alpha)\|_F \leq \|\Delta\|_F\) for any matrix \(\Delta\). This is a mild and well-justified assumption: the primary objective of post-hoc regularization is to enhance locality and mitigate interference, which is typically achieved by suppressing redundant components or projecting the perturbation onto a constrained subspace. Consequently, such operations naturally exhibit non-expansive behavior. This assumption holds for a wide range of prevalent regularizers, including RECT \citep{gu2024model}, PRUNE \citep{maperturbation}, AdaEdit \citep{li2025adaedit}, and SPHERE \citep{liu2026energyregularized}.

For history-agnostic and history-regularized editors,
post-hoc regularization is applied to each incremental
update. By the triangle inequality and the
non-expansiveness assumption, we have
\[
\|\mathbf{W}_t^{\mathrm{PR}}\|_F
\leq
\|\mathbf{W}_0\|_F+
\sum_{s=1}^{t}\|\Delta_s^{\mathrm{base}}\|_F.
\]
For history-agnostic editing, the per-step update satisfies
\[
\|\Delta_s^{\mathrm{HA}}\|_F
\leq
\frac{C_r}{2\sqrt{\lambda_{\min}}},
\]
which gives
\[
\|\mathbf{W}_t^{\mathrm{PR-HA}}\|_F
\leq
\|\mathbf{W}_0\|_F+
\frac{C_r}{2\sqrt{\lambda_{\min}}}t.
\]

For history-regularized editing, using
$\|A_s^{-1}\|_2\leq 1/(s\alpha_0)$, we obtain
\[
\|\Delta_s^{\mathrm{HR}}\|_F
\leq
\frac{C_rC_k}{s\alpha_0}.
\]
Consequently,
\[
\|\mathbf{W}_t^{\mathrm{PR-HR}}\|_F
\leq
\|\mathbf{W}_0\|_F+
\frac{C_rC_k}{\alpha_0}(1+\log t).
\]

For batch-equivalent editing, the
post-hoc regularization is applied to the batch-equivalent solution:
\[
\mathbf{W}_t^{\mathrm{PR-BE}}
=
\mathbf{W}_0+
\operatorname{PR}
\left(
\mathbf{W}_t^{\mathrm{BE}}-\mathbf{W}_0,\alpha
\right).
\]
By the non-expansiveness assumption, it follows that
\[
\begin{aligned}
\|\mathbf{W}_t^{\mathrm{PR-BE}}\|_F
&\leq \|\mathbf{W}_0\|_F+
\|\operatorname{PR}
(\mathbf{W}_t^{\mathrm{BE}}-\mathbf{W}_0,\alpha)\|_F\\
&\leq \|\mathbf{W}_0\|_F+
\|\mathbf{W}_t^{\mathrm{BE}}-\mathbf{W}_0\|_F\\
&\leq \|\mathbf{W}_0\|_F+
\frac{C_rC_k}{\alpha_0}.
\end{aligned}
\]
Therefore, post-hoc regularization preserves the
established weight growth upper bounds of all three
base editing methods, without changing their respective
growth rates.

This completes the proof.
\end{proof}

\subsection{Tightness of the Upper Bounds}
Consider the scalar setting with:
\begin{itemize}
    \item  $ K_0 = \sqrt{\lambda} $  for some  $ \lambda > 0 $ , so  $ K_0 K_0^\top = \lambda > 0 $ (we set $\lambda=1$);
    \item Edit inputs  $ k_t = 1 $  for all  $ t $  (hence  $ c_k = C_k = 1 $ );
    \item Fixed residual magnitude  $ |r^\star| = C_r > 0 $;
    
\end{itemize}

Then the history-agnostic update becomes
$$
W_t = W_{t-1} + \frac{r^\star}{1 + \lambda},
$$
which yields
$$
W_t = W_0+|r^\star| \sum_{s=1}^t \frac{1}{1 + \lambda} =W_0+ t \cdot \frac{C_r}{2} \lambda_{\min}(P_0)^{-1/2}.
$$
Here, we omit the symbol $\| \|$ for $W_t$ and $W_0$ as the weight at every editing step is a constant.

Similarly, the history-regularized update becomes
$$
W_t = W_{t-1} + \frac{r^\star}{t + \lambda},
$$
which yields
$$
W_t = W_0+|r^\star| \sum_{s=1}^t \frac{1}{s + \lambda} = W_0+ \frac{C_r C_k}{\alpha}(1 + \log t) (t\to \infty).
$$
And the batch-equivalent sequential update becomes
$$
W_t = W_{0} + t\cdot \frac{r^\star}{t + \lambda}=W_{0}+\frac{C_r C_k}{\alpha} (t\to \infty),
$$
This verifies the tightness of the upper bounds.

Finally, the upper bounds for post-hoc regularized
editing are also tight. This follows from the fact that
$\operatorname{PR}(\Delta,0)=\Delta$, under which
post-hoc regularization reduces to the corresponding
base editor. Since the upper bounds of the three base
editors are tight, the upper bounds established for
their post-hoc regularized variants are tight as well.
 
\section{Additional Experimental Setup}
\label{app:baseline_hahr}
\subsection{Evaluation Metrics}

We evaluate editing performance using three standard metrics \citep{liu2026energyregularized}. 
Let $f_{\theta}$ denote the original model and $f_{\theta^*}$ denote the model after sequential editing. For each edit instance, we are given:
\begin{itemize}
    \item The input prompt $x$ (e.g., "The president of the United States is"),
    \item The target output $y_{\text{target}}$ (e.g., "Joe Biden"),
    \item A set of paraphrased prompts $\mathcal{P}(x)$ that preserve the same meaning as $x$,
    \item A set of unrelated prompts $\mathcal{U}$ that are semantically independent of the edited fact.
\end{itemize}

Let $p_{\theta}(\cdot \mid x)$ denote the probability distribution over the output vocabulary produced by the model. The three metrics are formally defined as follows:

\textbf{Success Rate (Suc.)} measures whether the edited model assigns the highest probability to the target output $y_{\text{target}}$ for the original input $x$:
\[
\text{Suc} = \mathbb{E}_{(x, y_{\text{target}})} \left[ \mathbf{1}\left( \arg\max_y p_{\theta^*}(y \mid x) = y_{\text{target}} \right) \right],
\]
where the expectation is taken over all edit instances in the test set.

\textbf{Generality (Gen.)} measures whether the edited model generalizes the edited fact to paraphrased inputs:
\[
\text{Gen} = \mathbb{E}_{(x, y_{\text{target}})} \left[ \frac{1}{|\mathcal{P}(x)|} \sum_{x' \in \mathcal{P}(x)} \mathbf{1}\left( \arg\max_y p_{\theta^*}(y \mid x') = y_{\text{target}} \right) \right].
\]

\textbf{Locality (Loc.)} measures whether the editing process preserves the model's behavior on unrelated inputs. A good editor should maintain the original output distribution for unrelated prompts, i.e., the probability of the original output $y_{\text{original}}$ should remain unchanged from before editing. The metric is defined as:
\[
\text{Loc} = \mathbb{E}_{(x_{\text{unrelated}}, y_{\text{original}})} \left[ \mathbf{1}\left( \arg\max_y p_{\theta^*}(y \mid x_{\text{unrelated}}) = y_{\text{original}} \right) \right],
\]
A desirable editor should not degrade the model's performance on unrelated facts.
\subsection{Update Rules of Sequential LTE Baselines}
Below, we detail the update rules for the sequential LTE methods used as baselines in our study. For each method, we provide three variants: a history-agnostic (HA) form, which does not explicitly account for previous edits; a history-regularized (HR) form, which extends the regularization term to protect previously edited knowledge \citep{fangalphaedit,guotowards,fang2025hippocampal}; and a batch-equivalent form, which produces the batch solution at every editing step.

\paragraph{MEMIT-HA (History-Agnostic).}
The update rule is:
\begin{equation}
    W_{t} = W_{t-1} + r_t k_t^\top \big(\lambda_2\mathbf{K}_0 \mathbf{K}_0^\top + k_t k_t^\top \big)^{-1}.
\end{equation}

\paragraph{MEMIT-HR (History-Regularized).}
To protect previously edited knowledge, we extend the regularization term as follows:
\begin{equation}
    W_{t} = W_{t-1} + r_t k_t^\top \big(\lambda_2\mathbf{K}_0 \mathbf{K}_0^\top + \mathbf{K}_{[1:t]} \mathbf{K}_{[1:t]}^\top \big)^{-1}.
\end{equation}

\paragraph{MEMIT-BE (Batch-Equivalent).}
This method is introduced in Appendix~\ref{app:bese}. The update rule is:

\begin{align}
W_t &= W_{t-1} \left( \mathbf{I} - \lambda_1 k_t k_t^\top A_t^{-1} \right) + \lambda_1 \left( r_t + W_0 k_t \right) k_t^\top A_t^{-1}, \\
A_t &= \lambda_2 \mathbf{K}_0 \mathbf{K}_0^\top + \mathbf{K}_{[1:t]} \mathbf{K}_{[1:t]}^\top,
\end{align}

where we follow MEMIT's default settings with $\lambda_1 = 1$ and $\lambda_2 = 15000$ for fair comparison.

\paragraph{EMMET-HA (History-Agnostic).}
The update rule is:
\begin{equation}
    W_{t} = W_{t-1} + r_t \big(k_t^\top (\mathbf{K}_0 \mathbf{K}_0^\top)^{-1} k_t+\mu\mathbf{I}\big)^{-1} k_t^\top (\mathbf{K}_0 \mathbf{K}_0^\top)^{-1}.
\end{equation}
\paragraph{EMMET-HR (History-Regularized).}
We adapt the rule by updating the inverse covariance estimate to include prior edits:
\begin{equation}
    W_{t} = W_{t-1} + r_t \big(k_t^\top  (\mathbf{K}_0 \mathbf{K}_0^\top+ \mathbf{K}_{[1:t]} \mathbf{K}_{[1:t]}^\top)^{-1} k_t+\mu\mathbf{I}\big)^{-1}
    k_t^\top  (\mathbf{K}_0 \mathbf{K}_0^\top+ \mathbf{K}_{[1:t]} \mathbf{K}_{[1:t]}^\top)^{-1}.
\end{equation}

\paragraph{EMMET-BE (Batch-Equivalent).}
Following the procedure in Appendix~\ref{app:bese},
we derive the batch-equivalent sequential editing
algorithm for EMMET:
\begin{equation}
W_t=W_{t-1}
+\left(r_t-(W_{t-1}-W_0)k_t\right)k_t^\top C_t^{-1},
\end{equation}
where $C_t=\mu \mathbf{K}_0 \mathbf{K}_0^\top + \mathbf{K}_{[1:t]} \mathbf{K}_{[1:t]}^\top.$ This update is derived by reformulating the EMMET
batch solution using the matrix inversion identity,
which transforms the regularization term $\mu\mathbf{I}$
into $\mu \mathbf{K}_0 \mathbf{K}_0^\top$.  We follow EMMET's default setting with $\mu=0.1$
for a fair comparison.

\paragraph{AlphaEdit-HA (History-Agnostic).}
The update rule is:
\begin{equation}
    W_{t} = W_{t-1} +  r_t k_t^\top \mathbf{P} \big(\lambda_2\mathbf{I}  + k_t k_t^\top \mathbf{P} \big)^{-1},
\end{equation}
where \(\mathbf{P}\) is a projection matrix onto the null space of preserved knowledge.

\paragraph{AlphaEdit-HR (History-Regularized).}
To extend protection to previously edited knowledge, we include all past keys in the regularization term:
\begin{equation}
    W_{t} = W_{t-1} +  r_t k_t^\top \mathbf{P} \big(\lambda_2\mathbf{I} + \mathbf{K}_{[1:t-1]} \mathbf{K}_{[1:t-1]}^\top \mathbf{P} + k_t k_t^\top \mathbf{P} \big)^{-1}.
\end{equation}

\paragraph{AlphaEdit-BE (Batch-Equivalent).}
This method is introduced in Appendix~\ref{app:bese}. The update rule is:
\begin{align}
\mathbf{W}_{t} &= \mathbf{W}_{t-1} \big( \mathbf{I} - k_{t} k_{t}^\top \mathbf{P} A_{t}^{-1} \big) + \big( r_{t} + \mathbf{W}_0 k_{t} \big) k_{t}^\top \mathbf{P} A_{t}^{-1}, \\
A_t &= \lambda_2 \mathbf{I} + \mathbf{K}_{[1:t]} \mathbf{K}_{[1:t]}^\top\mathbf{P},
\end{align}
where we follow AlphaEdit's default settings with $\lambda_2 = 10$  for fair comparison.
\begin{figure}[t]
  \centering
  \includegraphics[width=1\linewidth]{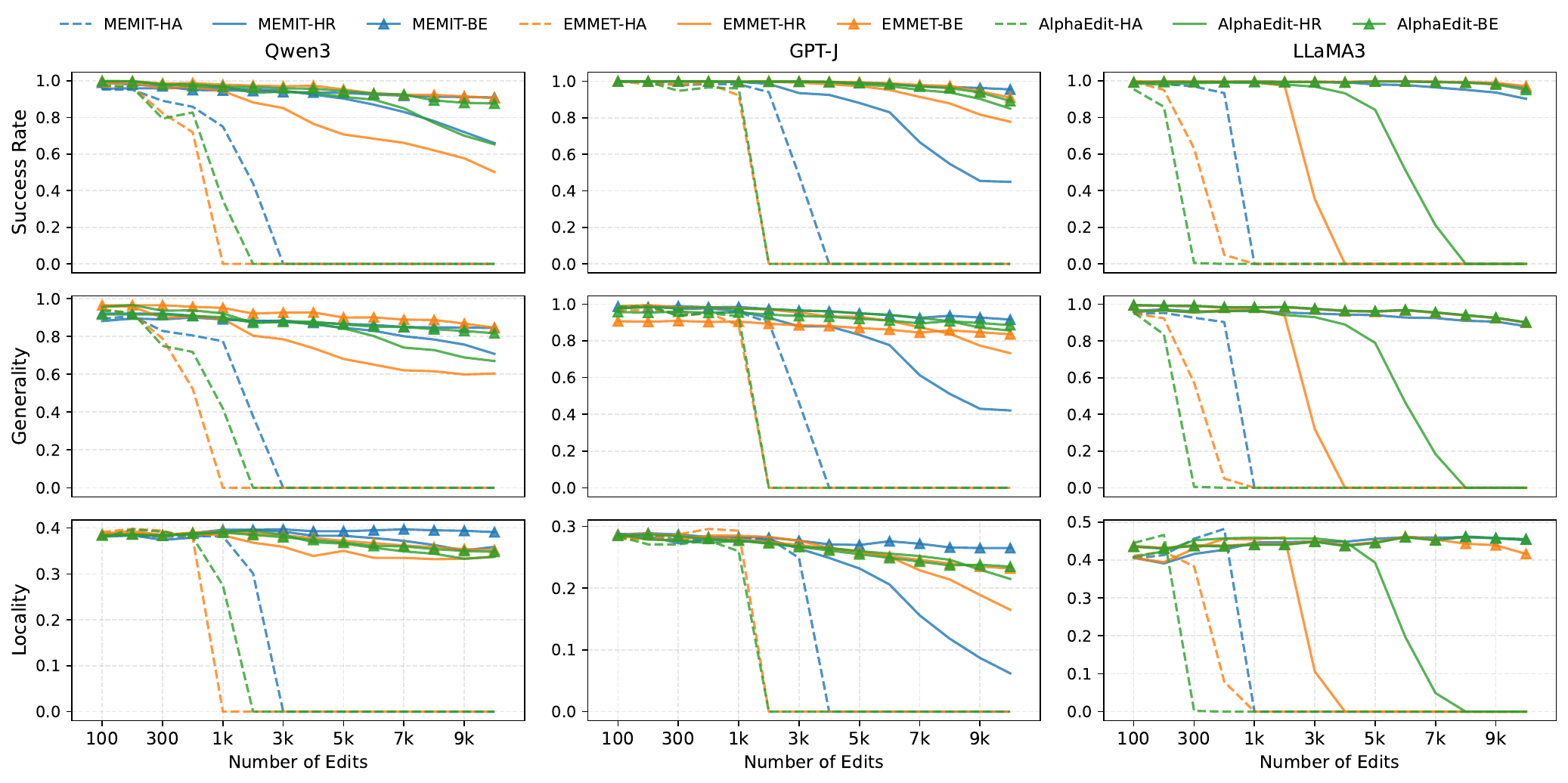}
\caption{
Editing performance (success rate, generality, locality) on the \textsc{ZsRE} dataset over 10,000 sequential edits. Columns (from left to right): Qwen3-8B, GPT-J, LLaMA3-8B-Instruct. Higher values indicate better performance for all metrics.}
  \label{fig:performance_zsre}
\end{figure}

\section{Additional Experimental Results}
\label{app:results}
\subsection{Additional Editing Results on \textsc{ZsRE}}
Figure~\ref{fig:performance_zsre} presents the editing performance of the three models on the \textsc{ZsRE} dataset over 10,000 sequential edits. Columns from left to right: Qwen3-8B, GPT-J, LLaMA3-8B-Instruct. Higher values indicate better performance for all metrics.

The results on \textsc{ZsRE} mirror the hierarchical performance pattern observed on \textsc{CounterFact}. First, history-regularized variants consistently outperform their history-agnostic counterparts, maintaining reliable performance over thousands of sequential edits. Second, our proposed batch-equivalent sequential editing method again achieves the best long-term performance, sustaining high accuracy across all three metrics even after tens of thousands of edits. These findings confirm that the observed performance hierarchy is robust across different factual editing benchmarks.

\subsection{Additional Stability Results}
\paragraph{Step-Per-Step Stability Results}
Figures~\ref{fig:sst_cf} and~\ref{fig:norm_cf} in the main paper report language understanding and parameter stability after subsampling every 400 steps to improve readability. Here we provide the complete step‐by‐step evolution of the weight norm.

\begin{figure}[t!]
\centering
\includegraphics[width=1\linewidth]{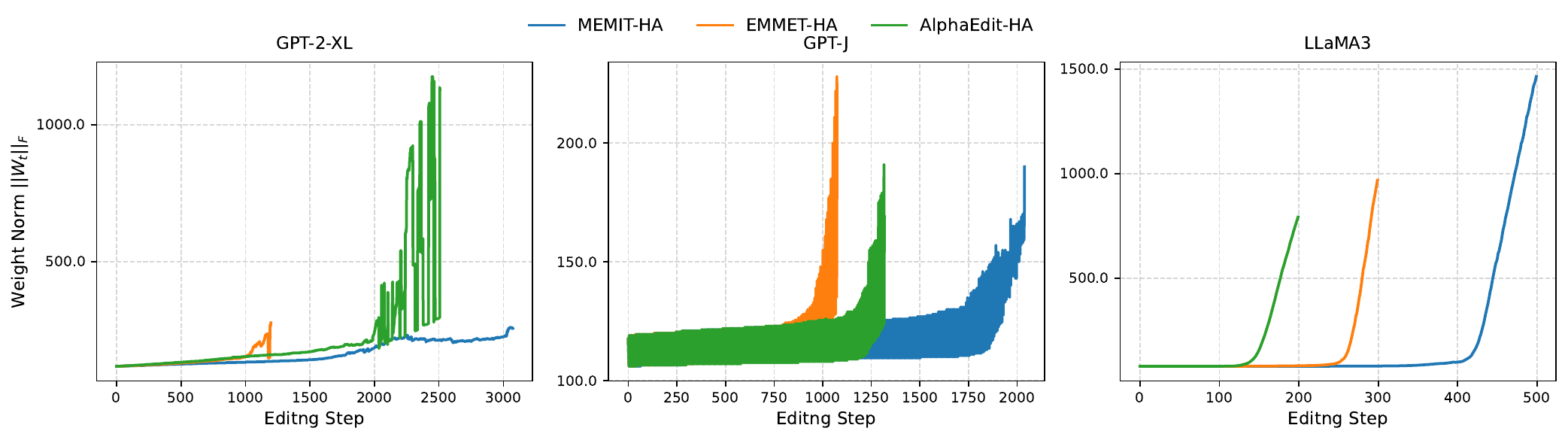}
\caption{Step‐by‐step evolution of the weight norm for GPT2-XL, GPT‑J and LLaMA3 during sequential editing on the \textsc{CounterFact} dataset.}
\label{fig:norm_ha_cf}
\end{figure}

\begin{figure}[t!]
\centering
\includegraphics[width=1\linewidth]{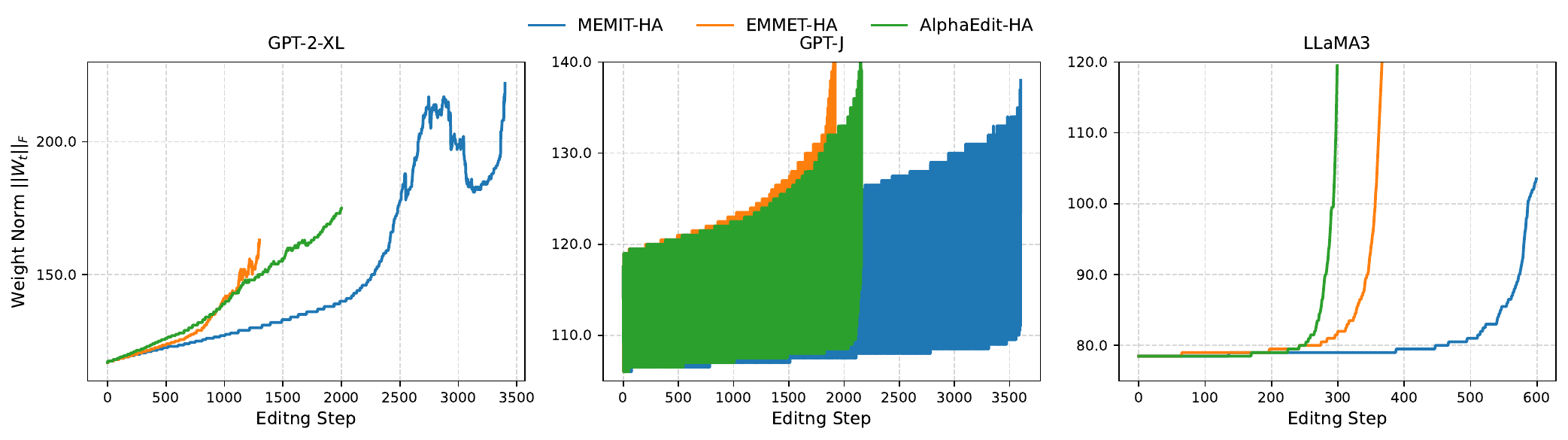}
\caption{Step‐by‐step evolution of the weight norm  for GPT2-XL, GPT‑J and LLaMA3 during sequential editing on the \textsc{ZsRE} dataset.}
\label{fig:norm_ha_zsre}
\end{figure}

Figures~\ref{fig:norm_ha_cf} and~\ref{fig:norm_ha_zsre} show the results on the \textsc{CounterFact} and \textsc{ZsRE} datasets, respectively, for three editing methods (EMMET‑HA, ALPHAEDIT‑HA, MEMIT‑HA) using GPT2-XL, GPT‑J and LLaMA3. As can be seen, the weight norm does not increase monotonically; it can drop substantially from one step to the next and exhibits strong oscillations, especially for GPT‑J. These oscillations motivate the subsampling strategy used in the main paper to demonstrate the overall trend of norm growth.

\begin{figure}[t!]
  \centering
  \includegraphics[width=1\linewidth]{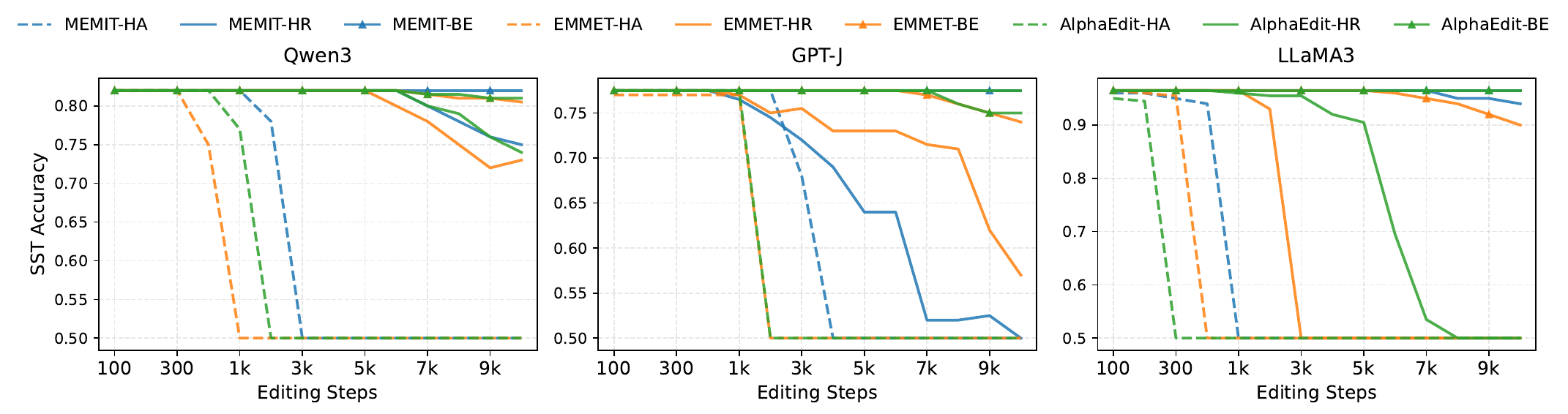}
\caption{
Evolution of language understanding capability (SST accuracy) during 10,000 sequential edits on the \textsc{ZsRE} dataset.
}
  \label{fig:sst_zsre}
\end{figure}

\begin{figure}[t!]
  \centering
  \includegraphics[width=1\linewidth]{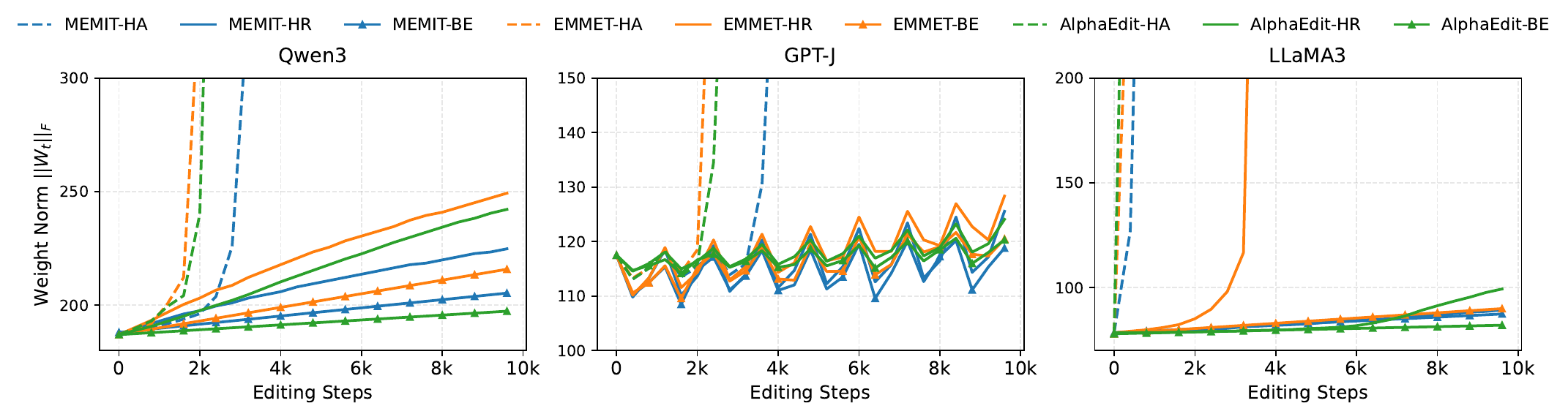}
\caption{Evolution of the weight norm during 10,000 sequential edits on the \textsc{ZsRE} dataset.}
  \label{fig:norm_zsre}
\end{figure}

\paragraph{Additional Stability Results on ZsRE} 
Figures~\ref{fig:sst_zsre} and~\ref{fig:norm_zsre} present the language understanding and parameter stability results on the \textsc{ZsRE} dataset, which mirror the hierarchical patterns observed on \textsc{CounterFact} (Figures~\ref{fig:sst_cf} and~\ref{fig:norm_cf}). History-agnostic editors exhibit rapid parameter drift and collapse of language understanding within a few hundred edits. History-regularized editors mitigate this drift but still struggle beyond thousands of edits. Batch-equivalent sequential editors achieve the highest stability, with \(\|\mathbf{W}_t\|_F\) remaining nearly constant and SST-2 accuracy staying close to its initial level over 10,000 edits.

\subsection{Additional Stability Metrics during Long-Term Editing}
\label{subsec:more_stability}

Beyond tracking the Frobenius norm of the updated weight $\|W_t\|_F$, we also evaluate two additional stability metrics:
\begin{itemize}
    \item \textbf{Weight Perturbation:} $\|\Delta_t\|_F = \|W_t - W_0\|_F$, measuring the total deviation from the original model.
    \item \textbf{Pretrained Knowledge Leakage:} $\|(W_t - W_0)\mathbf{K}_0\|_F$, quantifying how much the editing process affects preserved pretrained knowledge.
\end{itemize}

\begin{figure}[t!]
\centering
\includegraphics[width=1\linewidth]{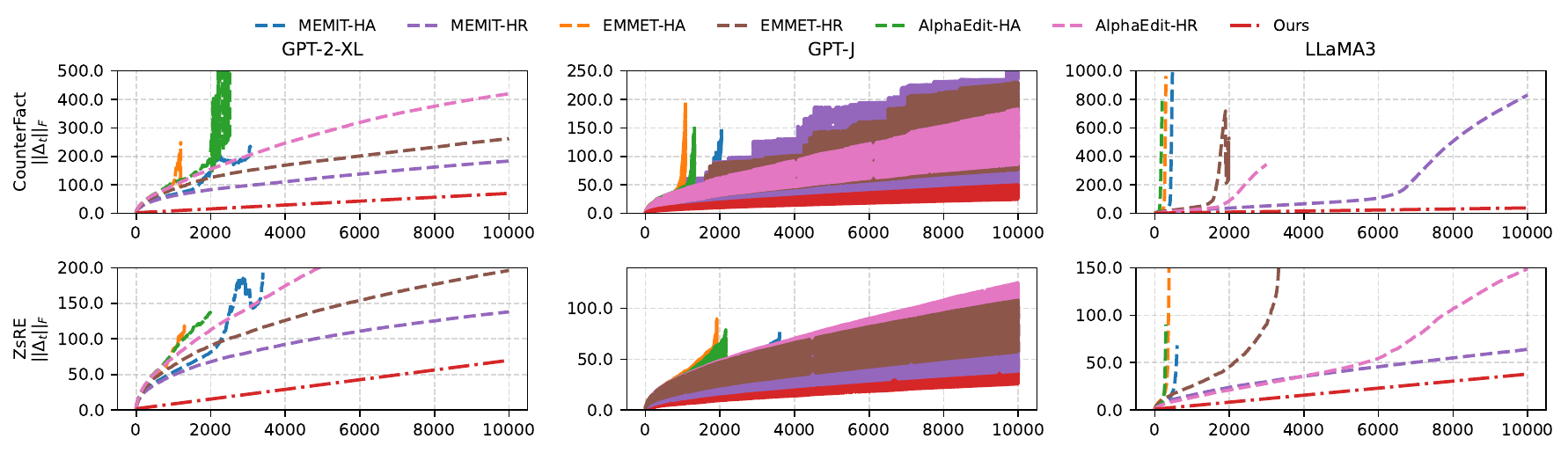}
\caption{Step‐by‐step evolution of the weight perturbation  for GPT2-XL, GPT‑J and LLaMA3 during 10,000 sequential edits.}
\label{fig:bese_delta_cf_zsre}
\end{figure}

\begin{figure}[t!]
\centering
\includegraphics[width=1\linewidth]{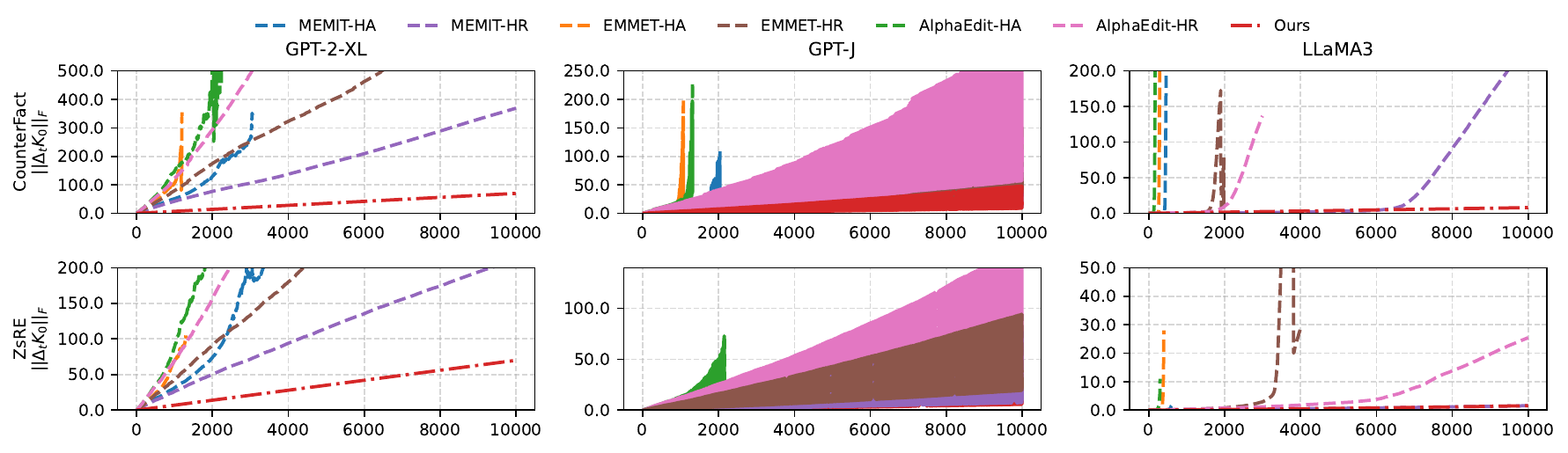}
\caption{Step‐by‐step evolution of the pretrained knowledge leakage for GPT2-XL, GPT‑J and LLaMA3  during 10,000 sequential edits.}
\label{fig:bese_deltak_cf_zsre}
\end{figure}

As shown in Figures \ref{fig:bese_delta_cf_zsre} and \ref{fig:bese_deltak_cf_zsre}, the plots (from left to right for GPT2-XL, GPT-J, and LLaMA3) confirm the same stability hierarchy across all models and datasets. The batch-equivalent sequential editor (MEMIT-BE) maintains the smallest weight perturbation and knowledge leakage, indicating the highest stability. In contrast, history-regularized and history-agnostic methods exhibit varying degrees of unbounded growth in these metrics over sequential edits.

\subsection{Additional Interpolation Results on \textsc{ZsRE}}

\begin{wrapfigure}[18]{r}{0.47\linewidth}
\centering
\includegraphics[width=\linewidth]{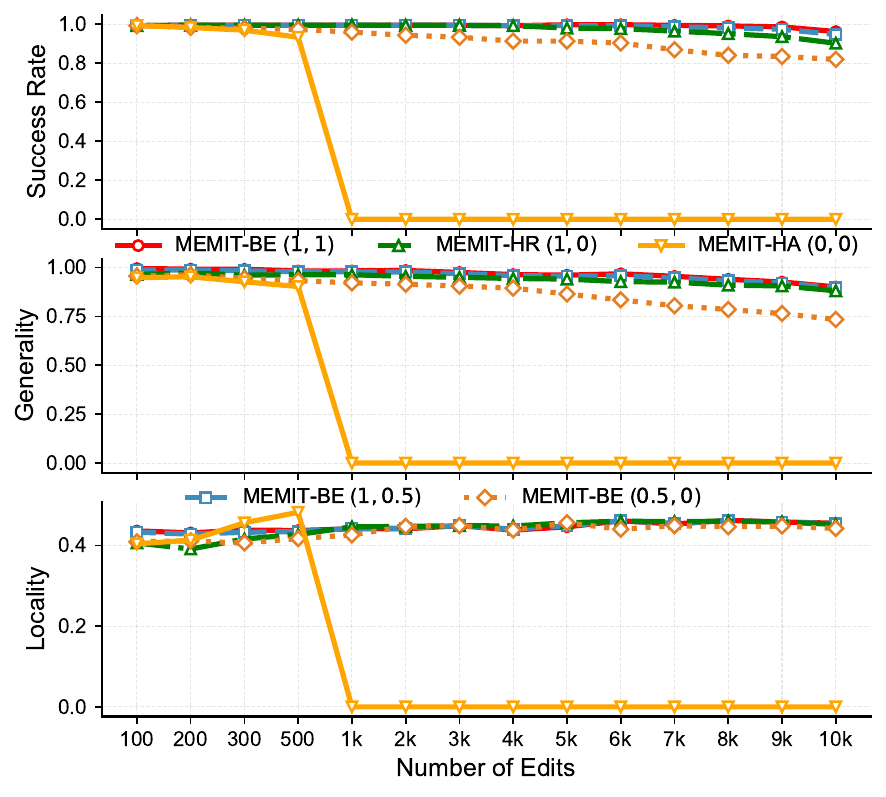}
\caption{Interpolation results on \textsc{ZsRE} using MEMIT on LLaMA3-8B-Instruct.}
\label{fig:memit_interpolation_llama_zsre}
\end{wrapfigure}

Figure~\ref{fig:memit_interpolation_llama_zsre} presents the interpolation results on the \textsc{ZsRE} dataset, mirroring the pattern observed on \textsc{CounterFact} (Figure~\ref{fig:memit_interpolation_llama_cf}). As $\beta_1$ increases from $0$ to $0.5$, the editing behavior smoothly transitions from history-agnostic to history-regularized, with corresponding improvements in long-term stability and performance (see MEMIT-BE(0.5,0)).




\subsection{Additional Results for Post-Hoc Regularized Editing}

\paragraph{Details about $\alpha$} 
The regularization strength $\alpha \in [0,1]$ controls the proportion of singular vectors corresponding to small singular values retained from the singular value decomposition of $W^\top W$, where $W$ is the weight matrix. Specifically, we decompose $W^\top W$ and sort the singular vectors in ascending order of their associated singular values (i.e., from smallest to largest). We then take the first $\alpha$ fraction of the singular vectors to form the matrix $U$. The projection matrix is constructed as $P = I - UU^\top$. When $\alpha = 0$, no singular vectors are retained ($U$ is empty), so $P = I$ and the original update is unchanged. When $\alpha = 1$, all singular vectors are retained, yielding a projection that maximally preserves locality. More details are referred to SPHERE \citep{liu2026energyregularized}.

\paragraph{Effect of post-hoc regularization on AlphaEdit-HR}
In the main paper, Figure~\ref{fig:post_reg_alphaedit_ha_llama_cf} demonstrates the effect of post-hoc regularization strategies (\textsc{RECT}, \textsc{PRUNE}, \textsc{AdaEdit}, \textsc{SPHERE}) applied to \textsc{AlphaEdit-HA} on the \textsc{CounterFact} dataset using LLaMA. To further examine the generality of these findings, we additionally apply the same post-hoc regularization strategies to \textsc{AlphaEdit-HR} under identical experimental conditions.

\begin{wrapfigure}[21]{r}{0.46\linewidth}
\centering
\includegraphics[width=\linewidth]{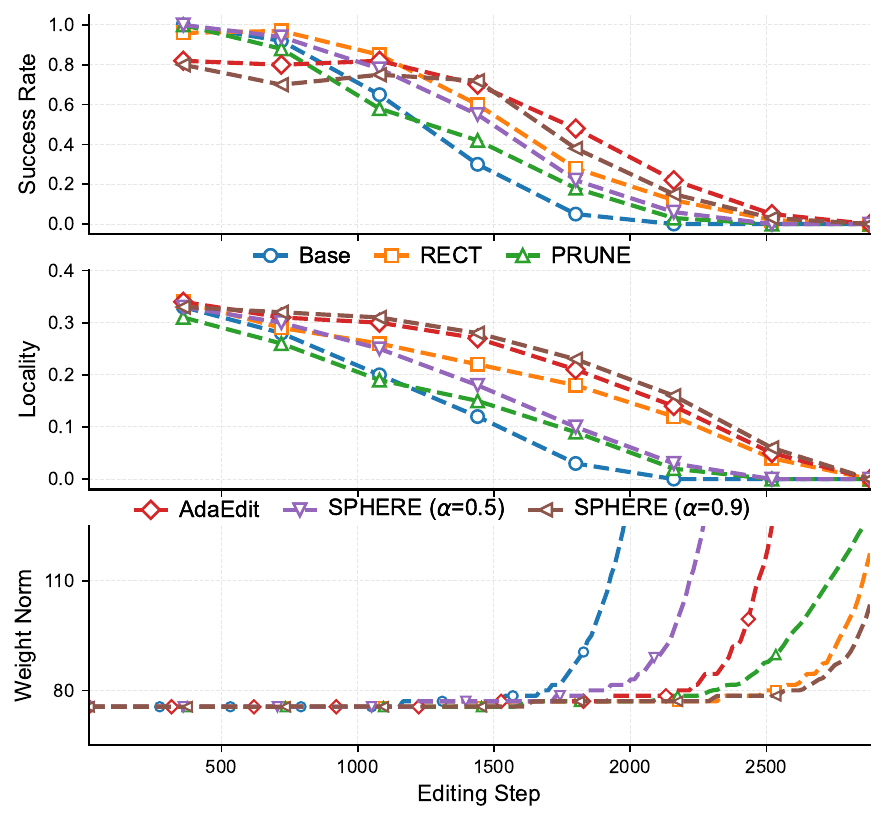}
\caption{Effect of post-hoc regularization on AlphaEdit-HR (base), with $\alpha$ denoting the regularization strength.}
\label{fig:post_reg_alphaedit_hr_llama_cf}
\end{wrapfigure}

Figure~\ref{fig:post_reg_alphaedit_hr_llama_cf} presents the editing performance and locality metrics for \textsc{AlphaEdit-HR} combined with \textsc{RECT}, \textsc{PRUNE}, \textsc{AdaEdit}, and \textsc{SPHERE}. 
Consistent with the observations for \textsc{AlphaEdit-HA},
post-hoc regularization applied to \textsc{AlphaEdit-HR}
exhibits similar behaviors. Specifically, increasing
$\alpha$ progressively improves locality by better
preserving the original weight norms and reducing
unintended drift in unrelated facts, but comes at the
cost of gradually declining editing success. Although
post-hoc regularization delays catastrophic failure,
allowing the model to withstand several hundred
sequential edits, its improvement over the unregularized
\textsc{AlphaEdit-HR} remains marginal compared with
\textsc{AlphaEdit-BE}.

\subsection{Extended Comparison with Additional Baselines}
\label{subsec:more_baselines}

We further compare our approach with two categories of existing knowledge editing methods: parameter-modifying methods and parameter-preserving methods.

\paragraph{Parameter-modifying methods.}
We include a broader set of sequential knowledge editing
baselines. For fine-tuning (FT) variants, we implement
LoRA-based fine-tuning \citep{hu2022lora} on the editing
layers used in locate-then-edit (LTE) methods, with a rank
of 8. At each editing step, a new LoRA module is initialized,
fine-tuned, and merged into the base model. Following LTE
practices, we augment inputs with randomly generated
prefixes. For hypernetwork-based methods, we include MEND
\citep{mitchellfast} and its reinforcement learning-enhanced
variant RLEdit \citep{lireinforced}, which train a hypernetwork
to generate weight updates from gradients. Among them,
RLEdit partially accounts for editing history through its
reward design. We also include PMET, an enhancement of
MEMIT that performs more precise interventions on hidden
representations within FFN modules \citep{li2024pmet};
SimIE, which approximates an ideal editor by projecting
vanilla updates to protect previously edited knowledge
\citep{guotowards}, using MEMIT as its base editor to
generate vanilla updates; and HSE, which incorporates the
covariance of previously edited keys to account for editing
history when optimizing the residual vector for the current
edit \citep{fang2025hippocampal}.

\begin{table*}[t!]
\centering
\caption{
Editing performance after 300 and 6,000 sequential edits on the \textsc{CounterFact} and \textsc{ZsRE} datasets using LLaMA3-8B-Instruct, GPT-2 XL, and GPT-J.
Higher values indicate better performance across all metrics: editing success rate (\textbf{Suc.}), generality (\textbf{Gen.}), and locality (\textbf{Loc.}).
}
\label{tab:results_add_modi}
\footnotesize
\setlength{\tabcolsep}{4pt}
\begin{tabular}{l  c c c c c c c c c c c c}
\toprule
\multirow{2}{*}{\textbf{Method}} & 
 \multicolumn{6}{c}{\textbf{CounterFact}} 
& \multicolumn{6}{c}{\textbf{ZsRE}} \\
\cmidrule(lr){2-7} \cmidrule(lr){8-13}
&\multicolumn{3}{c}{\textbf{$T=300$}}&\multicolumn{3}{c}{\textbf{$T=6000$}}&\multicolumn{3}{c}{\textbf{$T=300$}}&\multicolumn{3}{c}{\textbf{$T=6000$}} \\
\cmidrule(lr){2-4} \cmidrule(lr){5-7} \cmidrule(lr){8-10} \cmidrule(lr){11-13}
& \textbf{Suc.}& \textbf{Gen.} & \textbf{Loc.}
& \textbf{Suc.} & \textbf{Gen.} & \textbf{Loc.}
& \textbf{Suc.} & \textbf{Gen.} & \textbf{Loc.}
& \textbf{Suc.} & \textbf{Gen.} & \textbf{Loc.} \\
\midrule
\multicolumn{13}{c}{\textbf{LLaMA3-8B-Instruct}} \\
\midrule
FT & 
0.00 & 0.00 & 0.00 & 0.00 & 0.00 & 0.00
& 0.00 & 0.00 & 0.00 & 0.00 & 0.00 & 0.00 \\
MEND & 
0.00 & 0.00 & 0.00 & 0.00 & 0.00 & 0.00
& 0.00 & 0.00 & 0.00 & 0.00 & 0.00 & 0.00 \\
RLEdit & 
88.0 & 60.0 & 14.9 & 0.00 & 0.00 & 0.00
& 93.8 & 89.0 & 34.8 & 0.00 & 0.00 & 0.00 \\
PMET & 
79.7 & 61.3 & 20.4 & 0.00 & 0.00 & 0.00
& 96.8 & 92.5 & 36.4 & 0.00 & 0.00 & 0.00 \\
SimIE & 
98.8 & 68.0 & 20.1 & 0.00 & 0.00 & 0.00
& 96.5 & 92.0 & 36.0 & 74.8 & 61.3 & 27.4 \\
HSE & 
99.3 & 71.7 & 20.9 & 64.6 & 42.5 & 4.10
& 98.5 & 93.2 & 36.9 & 97.8 & 93.3 & 38.5 \\
\midrule
Ours & 
99.3 & 71.5 & 21.0 & 96.4 & 69.7 & 21.5
& 97.4 & 93.2 & 37.2 & 97.7 & 94.7 & 40.1 \\
\midrule
\multicolumn{13}{c}{\textbf{GPT-2 XL}} \\
\midrule
FT & 
0.00 & 0.00 & 0.00 & 0.00 & 0.00 & 0.00
& 0.00 & 0.00 & 0.00 & 0.00 & 0.00 & 0.00 \\
MEND & 
0.00 & 0.00 & 0.00 & 0.00 & 0.00 & 0.00
& 0.00 & 0.00 & 0.00 & 0.00 & 0.00 & 0.00 \\
RLEdit & 
85.2 & 58.3 & 5.10 & 0.00 & 0.00 & 0.00
& 92.4 & 87.6 & 23.5 & 0.00 & 0.00 & 0.00 \\
PMET & 
78.5 & 57.9 & 4.50 & 0.00 & 0.00 & 0.00
& 95.2 & 90.1 & 21.8 & 0.00 & 0.00 & 0.00 \\
SimIE & 
97.1 & 66.7 & 9.80 & 53.0 & 32.2 & 2.50
& 95.3 & 91.9 & 24.0 & 43.2 & 20.5 & 12.1 \\
HSE & 
98.8 & 69.0 & 10.1 & 77.0 & 48.2 & 5.80
& 96.5 & 92.0 & 24.9 & 74.8 & 61.3 & 15.5 \\
\midrule
ours & 
98.2 & 68.2 & 10.5 & 97.0 & 72.2 & 9.80
& 96.8 & 92.1 & 24.5 & 96.2 & 93.7 & 23.2 \\
\midrule
\multicolumn{13}{c}{\textbf{GPT-J}} \\
\midrule
FT & 
0.00 & 0.00 & 0.00 & 0.00 & 0.00 & 0.00
& 0.00 & 0.00 & 0.00 & 0.00 & 0.00 & 0.00 \\
MEND & 
0.00 & 0.00 & 0.00 & 0.00 & 0.00 & 0.00
& 0.00 & 0.00 & 0.00 & 0.00 & 0.00 & 0.00 \\
RLEdit & 
86.7 & 59.5 & 10.5 & 0.00 & 0.00 & 0.00
& 93.1 & 88.4 & 18.1 & 0.00 & 0.00 & 0.00 \\
PMET & 
79.1 & 60.7 & 9.80 & 0.00 & 0.00 & 0.00
& 96.0 & 91.3 & 21.0 & 0.00 & 0.00 & 0.00 \\
SimIE & 
97.5 & 67.2 & 12.3 & 48.7 & 23.8 & 3.20
& 95.8 & 91.5 & 26.1 & 74.1 & 61.0 & 18.2 \\
HSE & 
98.8 & 68.0 & 13.2 & 81.0 & 55.3 & 7.50
& 96.5 & 92.0 & 26.0 & 84.8 & 72.3 & 23.5 \\
\midrule
Ours & 
98.9 & 71.0 & 14.8 & 98.0 & 68.1 & 12.5
& 97.0 & 93.8 & 26.9 & 96.8 & 92.0 & 27.5 \\
\bottomrule
\end{tabular}
\end{table*}
Table~\ref{tab:results_add_modi} summarizes editing performance after 300 and 6,000 sequential edits on the CounterFact and ZsRE datasets using LLaMA3. The results show that most baselines struggle with long-term sequential editing, particularly at 6,000 edits. Methods not originally designed for sequential settings—such as FT, MEND, and PMET—degrade rapidly. RLEdit, which only accounts for editing history during the hypernetwork training phase, remains less effective in our strictly sequential setup (batch size = 1). SimIE falls short in long-term editing, as its history-aware protection mechanism tends to forget the current edit; notably, the original SimIE paper only evaluates performance under up to 1,000 sequential edits. HSE achieves better performance but still underperforms our method (MEMIT-BE), especially on the CounterFact dataset and with the GPT2-XL and GPT-J models.

\paragraph{Parameter-preserving methods.}
Beyond parameter-modifying methods, we further compare
our approach against methods that preserve the original
model parameters, including memory-based, adapter-based,
and in-context editors. SERAC stores edited facts in an
external memory module and learns a classifier to route
inputs between the memory and the base model
\citep{mitchell2022memory}. GRACE stores key-value 
pairs for edited facts and dynamically replaces forward
activations when a matching key is detected, avoiding
explicit parameter updates \citep{hartvigsen2023aging}.
WISE trains an auxiliary adapter to encapsulate edits and
selects the appropriate adapter during inference based on
input similarity \citep{wang2024wise}. IKE performs
in-context editing by prepending exemplar demonstrations
to the input, leveraging the model's in-context learning
capability without modifying its parameters
\citep{zheng2023can}. Finally, MEMOIR employs residual
memory and sparsifies input activations to assign edits
to distinct subsets of memory parameters, reducing
interference while supporting thousands of sequential
edits \citep{wangmemoir}.

Table~\ref{tab:results_pr} summarizes editing performance after 10,000 sequential edits on the CounterFact dataset across three model architectures. We report memory usage, per-edit inference time, success rate (Succ), generality (Gen), and locality (Loc).
Among the parameter-preserving baselines, we observe distinct trade-offs.

For locality, most parameter-preserving methods achieve strong locality. GRACE, SERAC, and IKE all maintain high locality across models (e.g., 21.6, 21.6, 21.1 on LLaMA3), as they do not modify the base model's parameters. WISE is the exception, showing noticeably lower locality (15.8 on LLaMA3) due to its trainable adapter module interfering with unrelated inputs.

For success rate, IKE and GRACE achieve near-perfect success on LLaMA3 (100.0 each), but GRACE's success comes at the cost of generality. SERAC fails entirely (0.0) across all models. WISE shows moderate success (97.2, 80.0, 68.9), decreasing with smaller model capacity. Notably, IKE shows poor performance on GPT2-XL and GPT-J as it relies heavily on the in-context learning capability of the base model.

For efficiency, our method (AlphaEdit-BE) achieves the lowest or near-lowest memory usage (36.0, 26.0, 8.0 GB) and per-edit time (0.32, 0.20, 0.07 s) across all three models. This efficiency stems from our method being a parameter-modifying approach, whereas parameter-preserving methods incur additional inference-time overhead due to their auxiliary memory or adapter modules.

Overall, our method strikes the best balance, delivering high editing efficacy and competitive locality with no additional inference cost.

\begin{table*}[t!]
\centering
\caption{Editing performance on the \textsc{CounterFact} dataset after 10,000 sequential edits. Memory and Time are reported per method.}
\label{tab:results_pr}
\footnotesize
\setlength{\tabcolsep}{4.5pt}
\begin{tabular}{l l c c c c c}
\toprule
\multicolumn{2}{c}{\textbf{Model \& Method}} & 
\textbf{Memory (GB) $\downarrow$} & 
\textbf{Time (s) $\downarrow$} & 
\textbf{Succ $\uparrow$} & 
\textbf{Gen $\uparrow$} & 
\textbf{Loc $\uparrow$} \\
\midrule
\multirow{6}{*}{LLaMA3} 
& IKE       & 46.0 & 1.28 & 100.0 & 51.2 & 21.1 \\
& SERAC     & 37.0& 0.62&  0.0  & 0.0  & 21.6 \\
& GRACE     & 36.0& 0.55&100.0  & 13.8 & 21.6 \\
& WISE      & 37.0& 0.38& 97.2  & 22.5 & 15.8 \\
& MEMOIR    & 39.0 & 0.47 & 78.1  & 40.2 & 21.5 \\
& Ours      & 36.0 & 0.32 & 92.5  & 71.4 & 20.8 \\
\midrule
\multirow{6}{*}{GPT-J} 
& IKE       & 37.0 & 0.71 & 75.3  & 32.5 & 12.0 \\
& SERAC     & 27.0& 0.45& 0.00 & 0.00& 12.5 \\
& GRACE     & 26.0& 0.33& 96.1 & 5.10& 12.0 \\
& WISE      & 29.0& 0.25& 80.0 & 46.7& 11.0 \\
& MEMOIR    & 30.0 & 0.28 & 85.5  & 40.4 & 12.2 \\
& Ours      & 26.0 & 0.20 & 95.6  & 61.5 & 11.9 \\
\midrule
\multirow{6}{*}{GPT2-XL} 
& IKE       & 11.0 & 0.32 & 61.0  & 23.1 & 10.1 \\
& SERAC     & 9.0& 0.17& 0.00 & 0.00& 10.2 \\
& GRACE     &  8.0& 0.16& 98.7 & 7.50& 10.6 \\
& WISE      & 8.0& 0.08& 68.9 & 25.0&  9.50 \\
& MEMOIR    &  9.0 & 0.14 & 76.3  & 29.6 & 10.4 \\
& Ours      &  8.0 & 0.07 & 90.2  & 59.4 &  9.20 \\
\bottomrule
\end{tabular}
\end{table*}

\section{Multi-Layer Editing with Cross-Layer Dependencies}
\label{app:multi}

Our batch-equivalence guarantee assumes that the key-residual
pairs defining the editing objective remain fixed throughout
the editing process. However, this assumption does not hold
for multi-layer editing algorithms with cross-layer dependencies. Specifically, updates to
earlier layers alter the representations propagated to
subsequent layers, thereby changing the keys and residuals
used to construct their editing objectives. Consequently,
the accumulated statistics in the sequential update may no
longer correspond to those of the batch solution.

Note that our main experiments adopt multi-layer editing with independent editing objectives across layers, using the editing objectives of MEMIT, EMMET, and AlphaEdit, but not using their multi-layer editing algorithms, which involve cross-layer dependencies. To investigate the impact of such dependencies, we consider two strategies for computing the desired values $\mathbf{v}$ (and the corresponding residuals) in MEMIT-style multi-layer editing:

\begin{itemize}
\item \textbf{Fixed-$\mathbf{v}$}: For each edit, we compute
the desired value $\mathbf{v}$ at the last editing layer
using the  \textbf{original} model weights $\mathbf{W}_0$.

\item \textbf{Dynamic-$\mathbf{v}$}: For each edit, we compute
the desired value $\mathbf{v}$ at the last editing layer
using the currently \textbf{updated} model, allowing it to reflect
changes introduced by preceding edits.
\end{itemize}

We evaluate the two strategies (\textbf{Fixed-$\mathbf{v}$} and
\textbf{Dynamic-$\mathbf{v}$}) using our batch-equivalent
sequential editing algorithm (Section~\ref{sec:method},
Appendix~\ref{app:bese} and Appendix \ref{app:precision_impl}) on \textsc{CounterFact} and
\textsc{ZSRE} with Qwen3-8B. We compare their results against
standard batch multi-layer MEMIT, which incorporates
cross-layer dependencies and computes the desired values
for each edit using the original model weights $\mathbf{W}_0$.
For both strategies, we adopt the direct batch-form
implementation described in Appendix~\ref{app:ncbe}.
Specifically, we incrementally maintain the matrices
$\mathbf{A}_j$ and $\mathbf{B}_j$ for each editing layer during sequential editing, and compute the updated
weights directly from the original model weights $\mathbf{W}_0$, rather than
recursively updating the weight matrix.  Table~\ref{tab:multilayer_dependency}
reports the results on 10,000 sequential edits.

\begin{table}[t]
\centering

\caption{Comparison of residual computation strategies when
applying our batch-equivalent sequential update to multi-layer
editing algorithms with cross-layer dependencies on Qwen3-8B.}

\label{tab:multilayer_dependency}
\setlength{\tabcolsep}{7pt}
\begin{tabular}{lccc}
\toprule
\textbf{Method}
& \textbf{SR.} $\uparrow$
& \textbf{Gen.} $\uparrow$
& \textbf{Loc.} $\uparrow$ \\
\midrule
\multicolumn{4}{c}{\textsc{CounterFact}} \\
\midrule
Batch
& 92.1 & \textbf{50.1} & 8.5 \\
Fixed-$\mathbf{v}$
& 89.7 & 48.2 & \textbf{9.0} \\
Dynamic-$\mathbf{v}$
& \textbf{93.8} & 48.1 & 8.3 \\
\midrule
\multicolumn{4}{c}{\textsc{ZSRE}} \\
\midrule
Batch
& 91.3 & 85.9 & 36.9 \\
Fixed-$\mathbf{v}$
& 88.8 & 83.7 & 37.4 \\
Dynamic-$\mathbf{v}$
& \textbf{98.3} & \textbf{89.9} & \textbf{37.8} \\
\bottomrule
\end{tabular}
\end{table}

Several observations emerge from the results. First,
Fixed-$\mathbf{v}$ underperforms batch editing in terms of
editing efficacy on both datasets. This suggests that,
under cross-layer dependencies and sequential editing, the editing statistics $\mathbf{A}$ and $\mathbf{B}$ 
accumulated for subsequent layers become outdated as
preceding layers are sequentially updated, causing the sequentially
maintained $\mathbf{A}$ and $\mathbf{B}$ matrices 
for non-first editing layers to deviate from those used in
batch editing.  Second, Dynamical-$\mathbf{v}$ yields editing improvements. On \textsc{CounterFact}, it achieves comparable success rate and generality to the batch baseline (93.8 vs. 92.1 and 48.1 vs. 50.1, respectively). On \textsc{ZSRE}, it substantially outperforms the batch baseline in both success rate (98.3 vs. 91.3) and generality (89.9 vs. 85.9), while maintaining comparable locality (37.8 vs. 36.9).

These results suggest that dynamically adapting the desired
values to the updated model better accommodates cross-layer
dependencies and can improve editing performance. In particular, although cross-layer dependencies break
the assumptions underlying our batch-equivalence guarantee,
dynamically computing the desired values provides a
practical means of accommodating these dependencies
throughout sequential editing, and can even yield better empirical performance
than batch editing.

\section{Future Work}
\label{app:futurework}
While current model editing techniques have made significant
progress in efficiently updating factual knowledge, several
fundamental challenges remain:

\begin{enumerate}

    \item \textbf{Multi-layer editing strategies.}
    Most existing methods rely on the residual propagation
    scheme introduced in MEMIT \citep{mengmass}, which has
    recently been shown to be suboptimal
\citep{li2024pmet,lirethinking,liubackward,liu2026cake}. Developing more effective
    layer-wise update mechanisms that better coordinate
    knowledge modifications across layers could further
    improve editing performance.

    \item \textbf{Fact localization in Transformers.}
    Evidence suggests that editing later components, such as
    FFNs and attention modules in deeper layers, can yield
    better results than modifying early FFN blocks
    \citep{yang2025fine,lireinforced,guomechanistic,zhou2025editing}.
    A systematic understanding of where factual knowledge
    resides and how it is distributed across layers and
    components could enable more precise and effective
    interventions.

    \item \textbf{Superficial editing.}
    Recent studies suggest that current parameter-modifying
    methods, including LTE-based approaches, often achieve
    only \textit{superficial editing} \citep{xie2025revealing}:
    they modify model outputs to match desired targets without
    necessarily internalizing the underlying knowledge.
    This limits their ability to support reasoning based on
    edited facts \citep{thedewikibigedit}. Enabling deeper
    integration of new knowledge and its compositional use
    in downstream reasoning remains an important challenge.

    \item \textbf{Beyond subject-centric editing.}
    Most existing approaches follow a subject-centric
    paradigm, anchoring edits to specific subject entities
    (e.g., ``Paris'' in ``Paris is the capital of France'').
    Although effective for isolated facts, this paradigm
    struggles with edits involving the same subject but
    different relations or contexts, particularly in
    long-form text
    \citep{jianganyedit,zeng2025docmedit,dong2025memit,xutanglescore}.
    Such scenarios can lead to interference, overwriting,
    and inconsistent updates. Developing relation-aware
    and context-sensitive editing frameworks is therefore
    important for handling more complex editing scenarios.

    \item \textbf{Coordination of editing targets.}
    Existing LTE methods typically optimize the desired
    residuals or target values of individual edits
    independently, without explicitly considering their
    mutual compatibility. This can create difficulties
    when multiple edits are jointly applied: independently
    specified targets may be difficult to reconcile in a
    single batch update, particularly when their
    representations overlap or interfere. A promising
    direction is to jointly optimize editing targets
    across an edit set, allowing individual targets to
    account for one another and produce more compatible
    key--value mappings before the weight update is computed
    \citep{dai2025eamet}.
\end{enumerate}
\section{Impact Statement}
While model editing enables efficient and localized updates to factual knowledge in large language models, it also introduces potential risks if deployed irresponsibly or maliciously. Editing techniques—including the method presented in this work—could be misused for targeted manipulation, such as injecting false information into specific factual associations without altering surface behaviors \citep{youssefposition,youssef2025tracing}. Moreover, sequential editing may accumulate stealthy changes that are difficult to detect or reverse once deployed. The present work focuses on the algorithmic foundations of sequential editing, and we explicitly disclaim its use for any harmful or unethical applications. We encourage the community to develop defensive mechanisms and ethical guidelines alongside editing capabilities.



\newpage
\section*{NeurIPS Paper Checklist}

The checklist is designed to encourage best practices for responsible machine learning research, addressing issues of reproducibility, transparency, research ethics, and societal impact. Do not remove the checklist: {\bf The papers not including the checklist will be desk rejected.} The checklist should follow the references and follow the (optional) supplemental material.  The checklist does NOT count towards the page
limit. 

Please read the checklist guidelines carefully for information on how to answer these questions. For each question in the checklist:
\begin{itemize}
    \item You should answer \answerYes{}, \answerNo{}, or \answerNA{}.
    \item \answerNA{} means either that the question is Not Applicable for that particular paper or the relevant information is Not Available.
    \item Please provide a short (1--2 sentence) justification right after your answer (even for \answerNA). 
\end{itemize}

{\bf The checklist answers are an integral part of your paper submission.} They are visible to the reviewers, area chairs, senior area chairs, and ethics reviewers. You will also be asked to include it (after eventual revisions) with the final version of your paper, and its final version will be published with the paper.

The reviewers of your paper will be asked to use the checklist as one of the factors in their evaluation. While \answerYes{} is generally preferable to \answerNo{}, it is perfectly acceptable to answer \answerNo{} provided a proper justification is given (e.g., error bars are not reported because it would be too computationally expensive'' or ``we were unable to find the license for the dataset we used''). In general, answering \answerNo{} or \answerNA{} is not grounds for rejection. While the questions are phrased in a binary way, we acknowledge that the true answer is often more nuanced, so please just use your best judgment and write a justification to elaborate. All supporting evidence can appear either in the main paper or the supplemental material, provided in appendix. If you answer \answerYes{} to a question, in the justification please point to the section(s) where related material for the question can be found.

IMPORTANT, please:
\begin{itemize}
    \item {\bf Delete this instruction block, but keep the section heading ``NeurIPS Paper Checklist"},
    \item  {\bf Keep the checklist subsection headings, questions/answers and guidelines below.}
    \item {\bf Do not modify the questions and only use the provided macros for your answers}.
\end{itemize}


\begin{enumerate}

\item {\bf Claims}
    \item[] Question: Do the main claims made in the abstract and introduction accurately reflect the paper's contributions and scope?
    \item[] Answer: \answerYes{} 
    \item[] Justification: The abstract and introduction clearly state the main claims of the paper.

\item {\bf Limitations}
    \item[] Question: Does the paper discuss the limitations of the work performed by the authors?
    \item[] Answer: \answerYes{} 
    \item[] Justification: They have been discussed in the Conclusion Section.

\item {\bf Theory assumptions and proofs}
    \item[] Question: For each theoretical result, does the paper provide the full set of assumptions and a complete (and correct) proof?
    \item[] Answer: \answerYes{} 
    \item[] Justification: They have been provided in the Appendix.

    \item {\bf Experimental result reproducibility}
    \item[] Question: Does the paper fully disclose all the information needed to reproduce the main experimental results of the paper to the extent that it affects the main claims and/or conclusions of the paper (regardless of whether the code and data are provided or not)?
    \item[] Answer: \answerYes{} 
    \item[] Justification: The hyperparameter setting is clearly stated in Appendix \ref{app:baseline_hahr}. 

\item {\bf Open access to data and code}
    \item[] Question: Does the paper provide open access to the data and code, with sufficient instructions to faithfully reproduce the main experimental results, as described in supplemental material?
    \item[] Answer: \answerYes{} 
    \item[] Justification: The data is publicly available and the detailed implementation algorithm for our method is provided in the Appendix. 

\item {\bf Experimental setting/details}
    \item[] Question: Does the paper specify all the training and test details (e.g., data splits, hyperparameters, how they were chosen, type of optimizer) necessary to understand the results?
    \item[] Answer: \answerYes{} 
    \item[] Justification: They are introduced in Section \ref{sec:exp}.

\item {\bf Experiment statistical significance}
    \item[] Question: Does the paper report error bars suitably and correctly defined or other appropriate information about the statistical significance of the experiments?
    \item[] Answer: \answerNo{} 
    \item[] Justification: The error bar is not important in our research field as the randomness is not significant.

\item {\bf Experiments compute resources}
    \item[] Question: For each experiment, does the paper provide sufficient information on the computer resources (type of compute workers, memory, time of execution) needed to reproduce the experiments?
    \item[] Answer: \answerYes{} 
    \item[] Justification: See Appendix.
    
\item {\bf Code of ethics}
    \item[] Question: Does the research conducted in the paper conform, in every respect, with the NeurIPS Code of Ethics \url{https://neurips.cc/public/EthicsGuidelines}?
    \item[] Answer: \answerYes{} 
    \item[] Justification: The research conforms to the NeurIPS Code of Ethics. 

\item {\bf Broader impacts}
    \item[] Question: Does the paper discuss both potential positive societal impacts and negative societal impacts of the work performed?
    \item[] Answer: \answerNA{} 
    \item[] Justification: This is foundational methodology research with no direct path to specific societal applications or harms.
    
\item {\bf Safeguards}
    \item[] Question: Does the paper describe safeguards that have been put in place for responsible release of data or models that have a high risk for misuse (e.g., pre-trained language models, image generators, or scraped datasets)?
    \item[] Answer: \answerNA{} 
    \item[] Justification: The paper does not release high-risk models (e.g., pretrained language models, image generators) or scraped datasets, so no safeguards are required.

\item {\bf Licenses for existing assets}
    \item[] Question: Are the creators or original owners of assets (e.g., code, data, models), used in the paper, properly credited and are the license and terms of use explicitly mentioned and properly respected?
    \item[] Answer: \answerYes{} 
    \item[] Justification: All existing assets used (e.g., datasets, code libraries) are properly cited in Section \ref{sec:exp} and Appendix.

\item {\bf New assets}
    \item[] Question: Are new assets introduced in the paper well documented and is the documentation provided alongside the assets?
    \item[] Answer: \answerNA{} 
    \item[] Justification: This paper does not introduce new assets.

\item {\bf Crowdsourcing and research with human subjects}
    \item[] Question: For crowdsourcing experiments and research with human subjects, does the paper include the full text of instructions given to participants and screenshots, if applicable, as well as details about compensation (if any)? 
    \item[] Answer: \answerNA{} 
    \item[] Justification: This paper does not involve crowdsourcing or research with human subjects.

\item {\bf Institutional review board (IRB) approvals or equivalent for research with human subjects}
    \item[] Question: Does the paper describe potential risks incurred by study participants, whether such risks were disclosed to the subjects, and whether Institutional Review Board (IRB) approvals (or an equivalent approval/review based on the requirements of your country or institution) were obtained?
    \item[] Answer: \answerNA{} 
    \item[] Justification: No human subjects research was conducted, so IRB approval is not applicable.

\item {\bf Declaration of LLM usage}
    \item[] Question: Does the paper describe the usage of LLMs if it is an important, original, or non-standard component of the core methods in this research? Note that if the LLM is used only for writing, editing, or formatting purposes and does \emph{not} impact the core methodology, scientific rigor, or originality of the research, declaration is not required.
    \item[] Answer: \answerNA{} 
    \item[] Justification: LLMs were not used as an important, original, or non-standard component of the core methodology. Any LLM usage was limited to writing/editing assistance and does not affect scientific rigor or originality.

\end{enumerate}

\end{document}